\documentclass[11pt]{article}
\usepackage[utf8]{inputenc}
\usepackage[legalpaper,margin=1.0in]{geometry}
\usepackage[round]{natbib}
\usepackage{microtype}
\usepackage{graphicx}
\usepackage{booktabs} 
\usepackage{cite}
\usepackage{amsmath,amssymb,xcolor}
\usepackage{amsfonts}
\usepackage{amsthm,comment}
 \usepackage{algorithm} 
\usepackage{algcompatible}
\usepackage[noend]{algpseudocode}
\usepackage{url}
\def\a{\mathbf{a}}
\def\b{\mathbf{b}}
\def\c{\mathbf{c}}
\def\d{\mathbf{d}}

\def\e{\mathbf{e}}
\def\y{\mathbf{y}}
\def\z{\mathbf{z}}
\def\A{\mathbf{A}}
\def\B{\mathbf{B}}
\def\C{\mathbf{C}}
\def\D{\mathbf{D}}
\def\Q{\mathbf{Q}}
\def\x{\mathbf{x}}
\def\y{\mathbf{y}}
\def\q{\mathbf{q}}
\def\I{\mathbf{I}}
\def\L{\mathcal{L}}
\def\X{\mathbf{X}}
\def\Y{\mathbf{Y}}
\def\R{\mathbf{R}}
\def\real{\mathcal{R}}
\def\W{\mathbf{W}}

\newcommand{\mr}{\mathrm}
\newcommand{\mb}{\mathbf}
\renewcommand{\le}{\leqslant}
\renewcommand{\ge}{\geqslant}
\DeclareMathOperator{\Tr}{Tr}

\renewcommand{\epsilon}{\varepsilon}

\DeclareMathOperator*{\argmin}{argmin}

\newtheorem{theorem}{Theorem}
\newtheorem{lemma}{Lemma}

\newtheorem{definition}{Definition}
\newtheorem{proposition}{Proposition}

\newcommand{\rvline}{\hspace*{-\arraycolsep}\vline\hspace*{-\arraycolsep}}
\author{Abiy Tasissa\thanks{Department of Mathematics, Tufts University, Medford, MA 02155, USA.} \and Pranay Tankala \thanks{School of Engineering and Applied Sciences, Harvard University, Cambridge, MA 02138, USA.}\and 
James M Murphy \footnotemark[1] \and Demba Ba\footnotemark[2]
}

\begin{document}

\title{K-Deep Simplex: Manifold Learning via Local Dictionaries}
\date{\vspace{-4ex}}

\maketitle

\begin{abstract}
We propose  \emph{$\mathrm{K}$-Deep Simplex (KDS)} which, given a set of data points, learns a dictionary comprising synthetic landmarks, along with representation coefficients supported on a simplex. KDS employs a local weighted $\ell_1$ penalty that encourages each data point to represent itself as a convex combination of nearby landmarks. We solve the proposed optimization program using alternating minimization and design an efficient, interpretable autoencoder using algorithm unrolling.  We theoretically analyze the proposed program by relating the weighted $\ell_1$ penalty in KDS to a weighted $\ell_0$ program. Assuming that the data are generated from a Delaunay triangulation, we prove the equivalence of the weighted $\ell_1$ and weighted $\ell_0$ programs. 
We further show the stability of the representation coefficients under mild geometrical assumptions. If the representation coefficients are fixed, we prove that the sub-problem of minimizing over the dictionary yields a unique solution. Further, we show that low-dimensional representations can be efficiently obtained from the covariance of the coefficient matrix. Experiments show that the algorithm is highly efficient and performs competitively on synthetic and real data sets.   

\end{abstract}

\section{Introduction}
\label{sec:intro}

Consider observations of the form $(\x_i,\y_i)_{i=1}^{n}$ with $\x_i \in \real^{m}$ and $\y_i\in \real^{d}$ denoting predictor and response variables respectively. We assume that $\y_i = f(\x_i)+ \epsilon_i$ where $\epsilon_i$ represents random i.i.d noise. A ubiquitous model is the standard linear regression which first posits that $f$ is linear and correspondingly estimates the model parameters via different methods (e.g., least squares). Rather than fixing a parametric model as in linear regression, non-parametric models learn the relation $f$ from the data with minimal assumption on $f$ (e.g., smoothness).  One popular class of non-parametric models is the local linear regression model \citep{loader2006local, stone1977consistent,cleveland1979robust,mclain1974drawing}. In contrast to linear regression which assumes a global form of $f$, local regression is based on approximating $f$ locally using linear functions. To be precise, the local linear fit at a point $\x_i$ is defined using a weight function $w_i$ that depends on distances to all other training data points (i.e., less weight is assigned to points far from $\x_i$).  Unlike the linear regression model where the global linear function is only needed for prediction at a test point, the locally linear model depends on the adaptive weight function and hence is non-parametric. One downside of this model is the curse of dimensionality where locality defined via distance functions implies that the weight function either considers nearly no neighbors or nearly all neighbors. To circumvent this limitation, dimensionality reduction-based approaches have been studied\citep{friedman1981projection, li1984classification}. Recent works have also explored local regression with new regularizations and recast it as an optimization problem over a suitably defined graph
\citep{hallac2015network,yamada2017localized,petrovich2020fast}. 

In this paper, we consider the unsupervised learning problem where we only have access to high-dimensional data  $(\y_i)_{i=1}^{n}$ with $\y_i\in \real^{d}$. This setting arises in many applications
and the raw high-dimensional representation presents challenges for computation, visualization, and analysis. The \emph{manifold hypothesis} posits that many high-dimensional datasets can be approximated by a low-dimensional manifold or mixture thereof. Hereafter, a $k$-dimensional submanifold $\mathcal{M}$ is a subset of $\real^{d}$ which locally is a flat $k$-dimensional Euclidean space \citep{lee2013smooth}. If the data lie on or near a \emph{linear} subspace, principal component analysis (PCA) can be used to obtain a low-dimensional representation. But, PCA may fail to preserve nonlinear structures. Nonlinear dimensionality reduction techniques  \citep{scholkopf1997kernel, tenenbaum2000global, roweis2000nonlinear, belkin2003laplacian, coifman2006diffusion} obtain low-dimensional representations while preserving local geometric structures of the data. 

Our main motivation is to develop a model akin to local linear regression in the unsupervised setting.  In fact, one of the critical parts of the local regression model is determining the neighborhood radius for each point such that the linear approximation is applied within the specified radius. We note that if the radius is set ``large", the linear approximation is sub-optimal. On the other hand, if the radius is set ``small", the locally linear estimate will be poor as it will only consider very few points. Given these extremes, determining the neighborhood radius, referred as the bandwidth function in the local regression literature \citep{loader2006local}, is of fundamental importance. A similar challenge also occurs in manifold learning algorithms in determining the number of neighbors (e.g., in locally linear embedding (LLE) \citep{roweis2000nonlinear}). 

Herein, to build our model for the unsupervised setting, we use synthetic points for the locally  linear approximation. To be precise, rather than considering the whole data set and considering neighboring points, we build local approximations by employing synthetic points that are to be learned. This approach resembles archetypal analysis \citep{cutler1994archetypal,van2019finding} where data points $(\y_i)_{i=1}^{n}$ with $\y_i\in \real^{d}$ are expressed as a convex combination of points $\a_1,\a_2,...,\a_m$ i.e., $\y_i = \sum_{j=1}^{m} x_j \a_j$ where $x_j\ge 0 \,\forall j$ and $\sum_{j=1}^{m} x_j=1$. The set of points $\{\a_i\}_{i=1}^{m}$ are known as the archetypes. In the original archetypal analysis paper \citep{cutler1994archetypal}, an alternating least squares problem is proposed to solve for the archetypes and the representation coefficients. To integrate archetypal analysis with local regression or manifold learning, we propose to represent each data point as a convex combination of archetypes with further regularization enforcing that more weight is assigned to nearby archetypes. One way to achieve this is by selecting a fixed number of nearby archetypes. While simple, estimating the optimal number of archetypes is challenging (as it inherently depends on the nonlinear structure of the data) and the resulting model is not flexible. Another way to impose locality is by enforcing that the weights are sparse for which the well-known $\ell_0$ minimization is a natural regularizer. 
Given that $\ell_0$ minimization is intractable, a widely adopted technique is based on its convex relaxation which yields the $\ell_1$ regularizer. However, since the weights are supported on the simplex, all the feasible solutions attain the same $\ell_1$ norm.

In this paper, we propose \emph{$\mathrm{K}$-Deep Simplex (KDS)},  a unified optimization framework for local archetypal learning. In KDS, each data point $\y\in \real^{d}$ is expressed as a sparse convex combination of $m$ atoms. These atoms define a dictionary $\A\in \real^{d\times m}$
to be learned from the data. To glean intrinsically low-dimensional manifold structure, we regularize to encourage representing a data point using nearby atoms. The proposed method learns a dictionary $\A$ and low-dimensional features with a structure imposed by convexity and locality of representation. To learn the atoms, we employ the alternating minimization framework which alternates between updating the atoms and updating the coefficients. The algorithm can also easily be mapped to a neural network architecture leading to interpretable neural networks. This mapping is along the lines
of algorithm unrolling~\citep{tolooshams2020deep,tolooshams2020convolutional,tolooshams2022stable,Monga2021}, an increasingly popular technique for structured deep learning.  
\subsection{Contributions}

This paper introduces a structured dictionary learning model based on the idea of representing data as a convex combination of local archetypes. One immediate advantage of the method is that it leads to an interpretable framework. Since the coefficients are non-negative and sum to 1, they automatically enjoy a probabilistic interpretation.

Another advantage of the proposed algorithm is its connection to structured compressed sensing. We show that the proposed locality regularizer can be interpreted as a weighted $\ell_1$ relaxation for a suitably defined $\ell_0$ minimization. Under a certain generative model of data, we show how the proposed weighted $\ell_1$ norm exactly recovers the underlying true sparse solution. In addition, for this generative model, we show stability of the weighted $\ell_1$ norm.  
In contrast to the standard compressed sensing setting which depends on coherence and the restricted isometry property (which do not hold in our setting), our analysis hinges on intrinsic geometric properties of data. 

The proposed locality regularizer is essentially a quadratic form of a Laplacian over a suitably defined graph. Since we learn a dictionary consisting of $m\ll n$ atoms, where $m$ is independent of $n$ and depends only on intrinsic geometric properties of the data, we show that the spectral embedding can be computed efficiently by only considering the $m \times m$ covariance matrix of the coefficient matrix. 

We discuss the alternating minimization framework to solve the main optimization problem. We argue that in the typical setting where $m\ll n$, the proposed algorithm is scalable. In addition, since our KDS embedding can be computed efficiently, this naturally leads to a scalable spectral clustering algorithm.  

We also map our iterative algorithm to a structured neural network. This mapping
is along the lines of iterative algorithm unrolling \citep{chang2019randnet, gregor2010learning, rolfe2013discriminative, tolooshams2018scalable,tolooshams2020deep,tolooshams2020convolutional,tolooshams2022stable,Monga2021} to solve our optimization problem. To be specific, we train a recurrent autoencoder with a nonlinearity that captures the constraint that our representation coefficients must lie on the probability simplex. To our knowledge, our use of algorithm unrolling for manifold learning is new.

\begin{figure*}[h!]
    \vskip 0.2in
    \centering
    \includegraphics[width=0.82\linewidth]{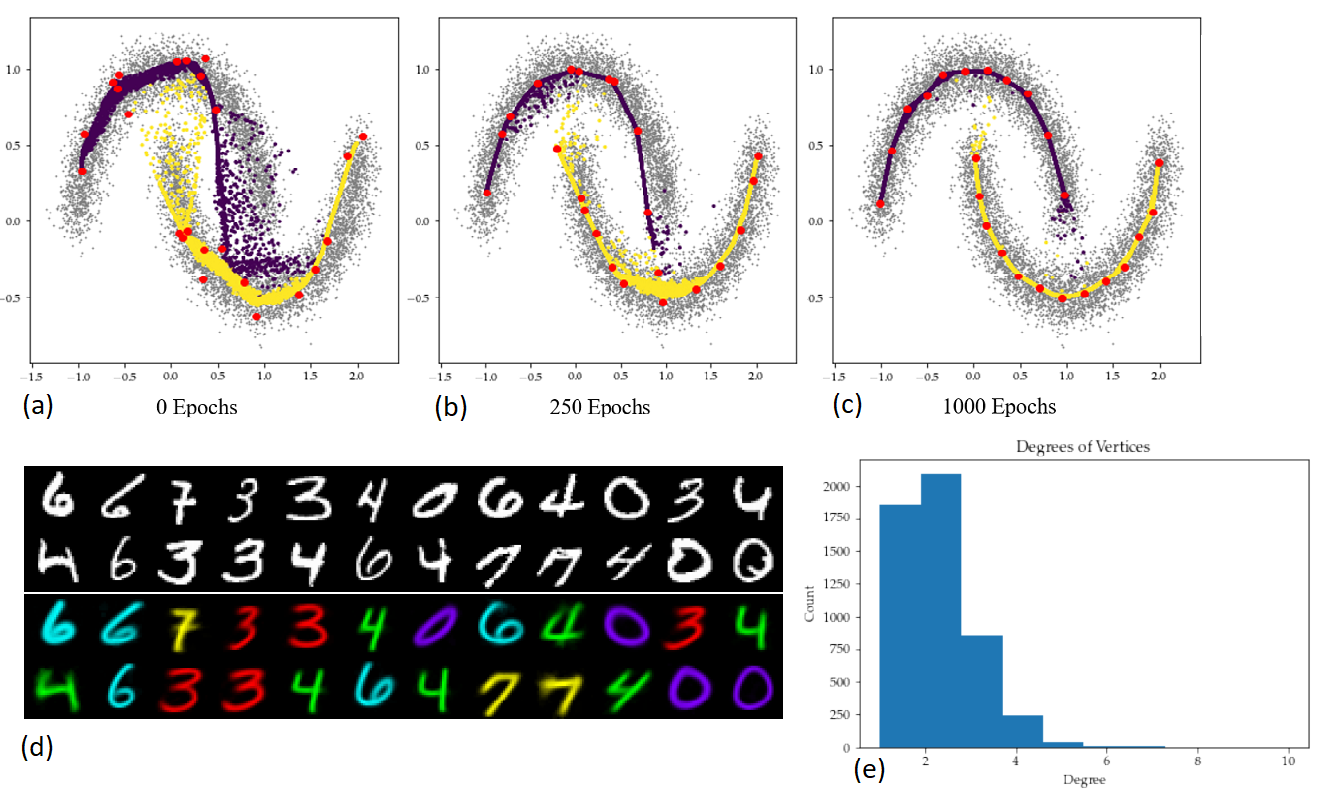}
    \caption{(a-c)  Training from a random initialization of atoms on the two moons data set.  (d) A subset of the randomly initialized atoms for MNIST-5 (digits 0, 3, 4, 6, 7) before training (black and white) and after training and clustering (color). The number of data points is $n\approx 35000$ and the number of atoms is $m=500$. (e) Degrees of vertices in the learned similarity graph. Despite being very sparse (most digits are represented using at most 5 atoms), the learned similarity graph retains enough information about the original data set that spectral clustering recovers these digits with 99\% accuracy.  }
       \label{fig:training_moon_mnist}
        \vskip -0.2in
\end{figure*}

For reproducibility, we will provide the code for all the experiments in this paper. To give a glimpse of the performance of KDS, Figure \ref{fig:training_moon_mnist} shows the atoms the autoencoder learns for the classic two moons dataset and digits from the MNIST-5 dataset (5 digits from the MNIST dataset).

\textbf{Differences from our prior work}: Previous work in \citep{tasissa_kds} by a subset of the authors of the present paper defines a weighted $\ell_0$ norm and shows that the weighted $\ell_1$ regularization studied in this paper recovers a unique solution under a certain generative model of data. 
Therein, we propose a simple alternating minimization algorithm to learn the sparse coefficients and the dictionary atoms and test it on two datasets. Some key differences between the work in \citep{tasissa_kds} and the current work are summarized below: 
\begin{enumerate}
\item Given fixed coefficients, we further consider the sub-problem of minimizing over the dictionary. Our result is summarized in Theorem \ref{thm:opt_over_atoms}.
\item The weighted $\ell_0$ norm defined in \citep{tasissa_kds} is a useful definition if the sparsity is fixed. If the sparsity is not fixed, Theorem 1 in \citep{tasissa_kds} is not correct and is not applicable. To fix this issue and have a theoretical result that does not depend on fixing the sparsity level, we define a more general weighted $\ell_0$ norm in this paper (see Definition 5).
\item We compare our method to more baselines and consider more datasets (e.g., images of faces, hyperspectral data).  
\item The main algorithm used in this paper is based on mapping the iterative algorithm to a neural network and departs from the previous algorithm which is based on alternating minimization.
\end{enumerate}
We also note that parts of the current work have appeared in our previously unpublished paper \citep{tankala2020k}. In contrast to these prior works, the current work presents new theory, comparisons to more baselines, a detailed review of related work, and interpretations of the proposed regularizer.
\subsection{Notation}

Lowercase and uppercase boldface letters denote column vectors and matrices, respectively. We denote the Euclidean, $\ell_0$, and $\ell_1$ norms of a vector $\x$, respectively as $||\x||_{2}$, $||\x||_{0}$ and $||\x||_{1}$. The Frobenius and operator norm of a matrix $\A$ are respectively denoted as $||\A||_F$ and $||\A||$. $\langle \x,\y \rangle$ denotes the Euclidean inner product. $\langle \mathbf{A}\,,\mathbf{B}\rangle$ denotes the trace inner product. The vector $\mathbf{1}$ denotes a vector whose entries are all $1$. $\Delta^{p}\equiv \{\mathbf{z} \in \real^{p}: \sum_{i=1}^{p} z_i = 1 , \mathbf{z}\ge \mathbf{0}\}$ denotes the probability simplex. Given a matrix $\A$, $\a_i$ denotes its $i$-th column. The set of $m\times n$ matrices where each column lies in the probability simplex $\Delta^m$ is denoted by $S$. $\text{diag}(\mathbf{x})$ represents a diagonal matrix whose entries are the vector $\x$. $\Tr(\A)$ denotes the trace of the matrix $\A$. The set of positive real numbers is denoted by $\real_{+}$. Given a scalar $x_i$, $\mathbf{1}_{\real_{+}}(x_i)$ denotes the indicator function whose value is $1$ if $x_i>0$ and is $0$ otherwise. $\e_j$ denotes a vector of zeros except a $1$ in the $j$-th position. $\sigma_{\max}(\A)$ and $\sigma_{\min}(\A)$ denote the largest and smallest singular values of $\A$. 

\section{Proposed Method: $K$-Deep Simplex}
\label{sec:method}

Let $\Y = [\y_1, \ldots, \y_n] \in \real^{d \times n}$ be a set of $n$ data points in $\real^d$. Our approach is to approximate each data point $\y_i$ by a convex combination of $m\ll n$ archetypes.  We define a dictionary $\A$ which is a collection of
the $m$ archetypes, $\A = [\a_1, \ldots, \a_m] \in \real^{d \times m}$. For sake of presentation, we first consider the case where the data points can be represented exactly as a convex combination of the archetypes. This leads to $\Y = \A \X$ where $\X = [\x_1, \ldots, \x_n] \in \real^{m \times n}$  is the coefficient or weight matrix. The convex combination implies $(\x_{i})_j\ge 0 \text{ for all } i \text{ and } j $ and $\X^\top\mb{1} = \mb{1}$. We note that this automatically provides us with a probabilistic interpretation of the coefficients. Next, we consider a suitable regularization with the aim that each data point is represented as a convex combination of its nearby archetypes. The regularization we consider is $\sum_{i,j} (\x_{i})_j\|\y_i - \a_j\|^2 $ where $(\x_{i})_j$ denotes the $j$-th entry of $\x_i$.
The resulting optimization program is given by

\begin{equation} \label{eq:optimization_alg}
\begin{aligned}
& \underset{\substack{\A \in \real^{d \times m} \\ \X \in \real^{m\times n}}}{\text{min}}
& & \sum_{i,j} (\x_{i})_j\|\mathbf{y}_i - \mathbf{a}_j\|^2 \\
& \text{subject to}
& & \Y = \A\X \\
& & & \sum_j (\x_{i})_j = 1 \quad \text{for } i = 1,2,...,n \\
& & & (\x_{i})_j \ge 0, \; \text{ for all } i,j.
\end{aligned}
\end{equation}

\subsection{KDS interpretations}
Below, we further explore the objective in the optimization program in \eqref{eq:optimization_alg} by discussing various interpretations. Note that, we focus on the locality regularization and do not consider the constraint $\Y = \A\X$. \\*

\textbf{Graph matching}: For a fixed $\A$, the objective in \eqref{eq:optimization_alg} can be related to graph matching. Consider a bipartite graph where the nodes are the data points and atoms. We consider matching the data points with the atoms using the coefficients to derive a cost matrix. Formally, we have the following 
\begin{align*}
\underset{\X \in S}{\min}\,\, \sum_{i,j}(\x_{i})_j\|\y_i - \a_j\|^2&= \underset{\X \in S}{\min}\,\,\Tr(\X^T\C)  = \underset{\X \in S}{\min} \,\,\langle \X\,,\C\rangle,
\end{align*}
where $\C \in \real_{+}^{m\times n}$ denotes a cost matrix defined as $C_{ij} =  ||\y_i - \a_j||_2^2$. The resulting problem is similar to the one to many graph matching problem \citep{cour2006balanced}. 

\textbf{Optimal transport}: Given the set of points $\Y$ and $\A$, we define empirical measures $\mu_y = \frac{1}{n}\sum_{i=1}^{} \delta_{\y_i}$ and $\mu_a = \frac{1}{m}\sum_{i=1}^{m} \delta_{\a_i}$ with $\delta$
denoting a Dirac measure.  The squared Wasserstein-2 distance between the probability measures $\mu_y$ and $\mu_a$ is defined as
\[
\mathcal{W}_2(\mu_y,\mu_a) = \underset{\gamma \in \Pi(\mu_y,\mu_a)}{\min} \sqrt{\sum_{i=1}^{n}\sum_{j=1}^{m} ||\y_i - \a_j||_2^2 \gamma_{ij}},
\]
where $\gamma$ is a joint probability measure over $\{\y_1,...\y_n\} \times \{\a_1,...,\a_m\}$ and $\Pi(\mu_y,\mu_a) = \{\gamma \in \real^{n\times m}|\gamma \mathbf{1}=\frac{1}{n}\mathbf{1}, \gamma^T\mathbf{1} =\frac{1}{m}\mathbf{1}\}$. 
If we let $\X = n\gamma^T$, the squared Wasserstein distance is equivalent to minimizing $\langle \X\,,\C\rangle$ over the set $\{\X \in \real_{+}^{n\times m}|\,\X^T\mathbf{1}=\mathbf{1}, \X\mathbf{1}=\frac{n}{m}\mathbf{1}\}$. In contrast to the standard regularizer which has a one-sided constraint (sum to 1 constraint as a result of convex combination), this new formulation further restricts the sum of coefficients
across rows placing a hard limit on how often a given atom is used to represent data points. 

\textbf{K-means}: Given data $\Y = [\y_1, \ldots, \y_n] \in \real^{d \times n}$, the K-means problem seeks to simultaneously find $m$ clusters with centers $\A=\{\a_1,...,\a_m\}$ and assign each data point to one of the $m$ clusters. The optimization problem is 
\begin{equation*}
\underset{\C\in \{0,1\}^{n\times m}, \{\a_i\}_{i=1}^{m}}{\min}\,\,\sum_{j=1}^{m} \sum_{i=1}^{n}    C_{ij} ||\y_i-\a_j||_2^2,
\end{equation*}
where $\C \in \{0,1\}^{n\times m}$ is a binary matrix satisfying $\forall i,\sum_{j=1}^{m} C_{ij}=1$. The above minimization problem resembles the objective in \eqref{eq:optimization_alg}. In fact, given $(\X^*, \A^*) =\underset{\A \in \real^{d \times m}\, \X \in S}{\arg\min} \sum_{i,j} (\x_{i})_j\|\y_i - \a_j\|^2 $, if each data point has a unique nearest atom in $\A^*$, it can be shown that each each column of $\X^*$ is one-sparse i.e., $\X^*$ is a binary assignment matrix. 

\textbf{Laplacian smoothness}: We first define a set of vertices by combining the data points and atoms. The coordinate representation of the combined vertices  is denoted by $\mathbf{R} = [\Y \,\,\, \A]\in \real^{d\times (n+m)}$. From this, we define a bipartite graph where edges only exist between data points and atoms i.e., in which an edge of weight $(\x_{i})_j$ connects the vertex $\y_i$ and the vertex $\a_j$. The weight matrix $\W \in \real^{(n+m)\times (n+m)}$ is 
\begin{equation}\label{eq:laplacian_weight}
\W = 
\begin{pmatrix}
  \mathbf{0}
  & \rvline & \X^T \\
\hline
  \X & \rvline &
  \mathbf{0}
\end{pmatrix}.
\end{equation}
The graph Laplacian is now defined as $\mathbf{L} = \mathbf{D}-\mathbf{W}$ where the diagonal degree matrix $\mathbf{D}  \in \real ^{(n+m)\times (n+m)} $ is defined as 
$D_{ii}  = \sum_{j=1}^{n+m} W_{ij}$. We now show how the locality regularizer is connected to the quadratic form of the Laplacian. 
\begin{proposition}
Let $\mathbf{R} = [\Y \,\,\, \A]\in \real^{d\times (n+m)}$. Then,
\[
\sum_{i=1}^{n} \sum_{j=1}^{m} (\x_{i})_j ||\y_i-\a_j||_2^2 = \Tr(\R\mathbf{L}\R^T).
\]
\end{proposition}
\begin{proof}
\begin{align*}
&\sum_{i=1}^{n} \sum_{j=1}^{m} (\x_{i})_j ||\y_i-\a_j||_2^2 \\ \nonumber
& = \sum_{i=1}^{n}\y_i^T\y_i
\sum_{j=1}^{m}(\x_{i})_j+ \sum_{j=1}^{n}\a_j^T\a_j\sum_{i=1}^{n}(\x_{i})_j-2\sum_{i,j}(\x_{i})_j\y_{i}^T\a_j\\ \nonumber
&= \sum_{i=1}^{n}\y_i^T\y_i
+\sum_{j=1}^{n}\a_j^T\a_j\, (\X\mathbf{1})_j-2\sum_{i,j}(\x_{i})_j\y_i^T\a_j\\ \nonumber
& = \Tr(\Y^T\Y\I)+\Tr(\A^T\A\text{diag}(\X\mathbf{1}))-\Tr(\R^T\R\W)\\
& = \Tr\left(
\begin{bmatrix}
\Y^T\Y & \Y^T\A\\
\A^T\Y & \A^T\A
\end{bmatrix}
    \begin{bmatrix}
    \mathbf{I}& \mathbf{0}\\
    \mathbf{0} & \text{diag}(\X\mathbf{1})    \end{bmatrix}\right)-\Tr(\R^T\R\W)\\
& =\Tr(\R^T\R\D)-\Tr(\R^T\R\W)\\
& = \Tr(\R^T\R(\D-\W)) = \Tr(\R\mathbf{L}\R^T)
\end{align*}
\end{proof}

Hence, the summation $\sum_{i=1}^{n} \sum_{j=1}^{m} (\x_{i})_j ||\y_i-\a_j||_2^2$ is precisely the \emph{Laplacian quadratic form} of the graph whose vertices are the data points and atoms where the weight function is the representation coefficients.

\section{Related Works}
One of the goals of the proposed model is to combine manifold learning with sparse coding/dictionary learning. To our knowledge, the first work that integrates sparse coding, manifold learning, and slow feature analysis is the sparse manifold transform framework proposed in \citep{chen2018sparse}. Therein, non-linear sparse coding using a learned dictionary is first used to map the data into a high-dimensional space. The next step extracts low-dimensional representations employing a matrix learned using a framework known as functional embedding \citep{chen2018sparse}. In this paper, the aim is a combination of linear sparse coding and dictionary learning. In addition, our analysis focuses 
on structured dictionaries coming from triangulation of a set of points. Below, we review related works in dictionary learning, manifold learning, and non-negative matrix factorization.

\subsection{Locality constrained dictionary learning}

Our work connects with sparse coding \citep{olshausen1996emergence} and dictionary learning. In sparse coding, given a fixed \emph{dictionary} $\A\in \real^{d\times m}$ of $m$ atoms, a data point $\y\in \real^{d}$ is represented as a linear combination of at most $k\ll m$ columns of $\A$. The dictionary $\A$ can be predefined \citep{bruckstein2009sparse} (e.g., Fourier bases, wavelets, curvelets) or adaptively learned from the data \citep{engan2000multi,aharon2006k,allard2012multi, maggioni2016multiscale}. The latter setting where the dictionary is simultaneously estimated with the sparse coefficients is the standard dictionary learning problem. We consider the prototypical form of the optimization objective for dictionary learning $\sum_{i=1}^{n}\frac{1}{2}||\y_i-\A\x_i||_2^2+R(\x_i,\A,\y_i)$ where $R(\x_i,\y_i,\A)$ is a regularization term on the representation coefficients, the dictionary atoms and the data points.

In Table \ref{tab:locality_constrained_review}, we review related works in graph regularized coding and locality constrained coding. The main idea in these works is to employ a Laplacian smoothness regularization such that if two data points are close, the regularization encourages their coefficients to be similar \citep{dornaika2019sparse,cai2010graph}. A few remarks are in order in how KDS compares to these methods. First, in KDS regularization, the underlying graph is not fixed but iteratively updated since the weights of the graph depend on the sparse representation coefficients. This is in contrast to methods that consider $\text{trace}(\X\L\X^T)$ where $\L$ is a priori fixed based on similarity of the data points. In Table \ref{tab:locality_constrained_review}, the closest methods to KDS are \citep{hu2014smooth,yu2009nonlinear,wang2010locality,elhamifar2011sparse}. However, the coefficients in these methods do not lie on the simplex and the regularizers are based on $(\x_{i})_j^2$ or $|(\x_{i})_j|$. The implication of these choices is that the sparse coding step in \citep{hu2014smooth,wang2010locality} yields a unique solution. This departs from our setup where  
the sparse coding step in general does not have a unique solution. In addition, the aforementioned works lack theoretical analysis that shows that the sparse coding step provably results a sparse solution. The sparse manifold clustering and embedding  algorithm (SMCE) \citep{elhamifar2011sparse} employs proximity regularization that promotes representation using local dictionaries. A drawback of SMCE is its computational inefficiency since the dictionary is essentially all the data points. Focusing on the problem of clustering, the work in \citep{ding2023unsupervised} introduces an optimization framework aimed at jointly learning a union-of-subspace representation and performing clustering. In this manuscript, the optimization objective retains a broad scope, learning representations that are not tailored to a specific end task. 
Finally, the work in \citep{zhong2020subspace} proposes a similar regularization to ours with the authors referring to it as ``adaptive distance regularization''. However, the methodology therein is based on using the data matrix as a dictionary and lacks theoretical analysis. Finally, we refer the reader to \citep{abdolali2021beyond} to find a comprehensive overview of nonlinear manifold clustering algorithms.

\begin{table}[!t]
\caption{related work\label{tab:locality_constrained_review}}
\centering
\begin{tabular}{l|l|ll}
Work & $R(\X,\Y,\A)$ & Notes on constraints \\ \hline 
\citep{zheng2010graph} & $\text{trace}(\X\L\X^T)+\lambda ||\X||_{1}$ & Sparse $\X$ and $||\a_i||_2^2\le c$\\
    &  &  $\L$ priori fixed    \\\hline
\citep{huang2015new}    & None & Simplex constraints on $X$ \\ \hline
\citep{hu2014smooth} & $\text{trace}(\X\L\X^T)$ &  $\L$ priori fixed \\\hline 
    
\citep{wang2010locality}   &  $\sum_{i,j} (\x_{i})_j^2 \exp\left(\frac{||\y_i-\a_j||}{\sigma}\right)$ & $\X^T\mathbf{1}=\mathbf{1}$ and $||\a_i||_2^2\le c$ \\\hline 
\citep{lcdl}   &  $\sum_{i,j} (\x_{i})_j^2 ||\y_i-\a_j||^2+\lambda||\X||_F^2$ & $\X^T\mathbf{1}=\mathbf{1}$, $(\x_{i})_j$ set to zero \\ 
  &   & based on neighborhood  \\ \hline 
\citep{jiang2021locality}   & $\text{trace}(\X\L\X^T)+\lambda ||\X||_{0}$ & Sparse $\X$, $\L$ priori fixed\\ \hline
  \citep{yu2009nonlinear} & $\sum_{i,j} |(\x_{i})_j|\,||\y_i-\a_j||^{1+p}$   & $\X^T\mathbf{1}=\mathbf{1}$. \\ \hline
  \citep{lcdl_support_vector}& $\text{trace}(\X^T\L\X)$   & $||\a_i||^2=1$. An additional   \\ 
     & &  SVM regularization \\\hline
\citep{zhong2020subspace} & $\sum_{i,j} (\x_{i})_j\,||\y_i-\y_j||^{2}+||\X||_F^2$  & Simplex constraints, \\
& $\text{diag}(\X)=\mathbf{0}$   & No dictionary learning \\ \hline
\citep{elhamifar2011sparse} & 
$\sum_{i,j} Q_{ij} (\x_{i})_j$ & $\X^T\mathbf{1} = \mathbf{1}$\\
 & & $\mathbf{Q} \equiv$ proximity regularizer \\\hline
\end{tabular}
\end{table}

\subsection{Manifold learning}
Our setup is along the lines of methods that learn local or global features of data using neighborhood analysis. For instance, locally linear embedding (LLE) \citep{roweis2000nonlinear} provides a low dimensional embedding using weights that are defined as the reconstruction coefficients of data points from their neighbors. The choice of the optimal neighborhood size is important for LLE as it determines the features obtained and subsequently the performance of downstream tasks. Geometric multiresolution analysis (GMRA)
is a fast and efficient algorithm that learns multiscale representations of the data based on local tangent space estimations \citep{allard2012multi, maggioni2016multiscale}. Since the dictionary elements used to reconstruct are defined locally, GMRA is not immediately useful for global downstream tasks, e.g., clustering. We also note that the work in \citep{multiscale_regression} develops a theoretical framework for regression on low-dimensional sets embedded in high dimensions. The regression is done via local polynomial fitting which resembles local convex approximation in KDS albeit the former method is applied to the supervised setting. 
  
\subsection{Scalable manifold learning via landmarks}

For large datasets, the embedding step in manifold learning techniques which typically involves a spectral problem can be costly. One approach to circumvent the computational challenge is based on finding an approximate solution by first identifying a subset of points designated as landmarks or exemplars. For instance, the works in 
\citep{silva2002global,de2004sparse} propose landmark isometric feature mapping (Isomap) and landmark multidimensional scaling (MDS) which are respectively scalable versions of Isomap \citep{tenenbaum2000global} and classical MDS\citep{torgerson1952multidimensional,gower1966some,young1938discussion}. The work in \citep{chen2011large} first considers sparse coding (assuming pre-computed $m$ landmarks) of all data points to obtain a sparse representation matrix $\mathbf{Z}\in \real^{m\times n}$. It then obtains spectral embeddings using the right singular vectors of a scaled $\mathbf{Z}$. Another approach along the lines of our work is the work in \citep{vladymyrov2013locally} which proposes an efficient version of the locally linear embedding method using landmarks. In contrast to our approach which learns the landmarks, we note that the methods in \citep{vladymyrov2013locally,chen2011large} identify the landmarks from the full data using strategies such as random sampling and clustering. A method inspired by LLE for semi-supervised learning, local anchor embedding (LAE), is proposed in \citep{lae_algo}. In this approach, the anchors are centers learned from the K-means algorithm. To obtain the representation coefficient of each data point, LAE solves a least squares problem in a dictionary of $s$-nearest anchors and with coefficients restricted on the simplex. Compared to our approach, the anchor learning step is disjoint from the sparse coding step in LAE. In addition, while LAE introduces sparsity by setting number of nearest anchors, our approach is based on promoting sparsity via a flexible proximity regularization. There are scalable landmark/exemplar methods for sparse subspace clustering e.g., \citep{you2018scalable,abdolali2019scalable,matsushima2019selective} but subspace clustering stipulates global affine structure that is not directly applicable to the general case of nonlinear manifolds.

\subsection{Non-negative matrix factorization}
Non-negative matrix factorization (NMF) considers the problem of approximating a non-negative data matrix using underlying components that are also non-negative\citep{lee1999learning,gillis2020nonnegative}. 
Let $\real^{m\times n}_{\ge 0}$ denote the set of $m\times n$ non-negative matrices. Given a data matrix $\Y \in \real^{d\times n}_{\ge 0}$, approximate NMF seeks non-negative matrices $\mathbf{W}\in \real^{d\times m}_{\ge 0}$ and $\mathbf{H}\in \real ^{m\times n}_{\ge 0}$ that best approximate the data. Choosing the Euclidean distance as a loss function, the problem can be formulated as $\underset{\mathbf{W}\in \real^{d\times m}_{\ge 0},\mathbf{H}\in \real ^{m\times n}_{\ge 0}}{\min}\,||\Y-\mathbf{W}\mathbf{H}||_F^2$. Different models on NMF put forth various conditions on the data matrix and the components. The work in 
\citep{ding2008convex} proposes a convex-model for NMF for a general data matrix with the restriction that $\mathbf{H}$ is non-negative and the columns of $\mathbf{W}$ lie in the column space of $\Y$. A similar work to ours is in \citep{lin2018maximum} where the authors propose simplex structured matrix factorization (SSMF) which considers the recovery of $\mathbf{W}$ and $\mathbf{H}$ given a generic data matrix with the restriction that $\mathbf{H} \in S$. Therein, the authors show that the exact $\mathbf{W}$ can be recovered by considering a maximum volume ellipsoid inscribed in the convex hull of the data points. We note that the model assumption in \citep{lin2018maximum} assumes a full column rank $\A$ and a full row rank $\X$ which we do not assume in our setting. Further discussion of different assumptions for identifiability of SSMF can be found in \citep{abdolali2021simplex}. Finally, the works in \citep{greer2011sparse} and \citep{charles2011learning} in hyperspectral imagery study a similar problem as ours but with the difference that the former considers a non-negativity constraint and the latter uses the $\ell_0$ regularizer on the simplex.   

\section{Theoretical analysis}

To solve the optimization program in \eqref{eq:optimization_alg}, a common approach is alternating minimization which is comprised of two steps. The first step is sparse coding and the second step is dictionary learning. 
In this section, we provide theoretical analysis for the sparse coding and dictionary learning steps of our proposed optimization program in \eqref{eq:optimization_alg}. The sparse coding problem fixes $\A$ 
and optimizes over $\X$ while the dictionary learning problem fixes $\X$ and optimizes for $\A$. We also discuss how to obtain a low-dimensional embedding of data points. Part of this analysis was completed in our prior work in \citep{tasissa_kds}. 

\subsection{Sparse coding}
The theoretical analysis for the sparse coding step assumes a specific model for the atoms and for generating the data points. Before describing the model, we start with 
essential background information on $d$-simplices, triangulations and a Delaunay triangulation. 

\begin{definition}
A $d$-simplex is the convex hull of a set of $d+1$ points 
$\{\a_0,\a_1,..,\a_d\}$  in $\real^d$. 
\end{definition}
For example, a 0-simplex and 1-simplex respectively correspond to a point and a line segment. The $d+1$ points that determine the $d$-simplex are called vertices of the simplex. Next, we define the $s$-face of a $d$-simplex. The definition is restated from \citep{cignoni1998dewall}.
\begin{definition}
    An s-face of a simplex is the convex combination of a subset of $s+1$ vertices of the simplex.
\end{definition}
For example, a $0$-face corresponds to a point, a $1$-face is an edge and a $2$-face is a triangular facet. The next definition concerns triangulation given a set of points. For the purposes of our analysis, we use the following definition \citep{chen2004optimal,cignoni1998dewall}. 
\begin{definition}
Given a set of points $\mathbf{P}=\{\mathbf{p}_1,\mathbf{p}_2,...,\mathbf{p}_m\}$ in $\real^{d}$, a triangulation $T$ is a set of $d$-simplices that partition the convex hull of $\mathbf{P}$ such that the intersection of any two simplices in $T$ is either empty or a common face.    
\end{definition}
We now proceed to define the main object of our theoretical analysis, the Delaunay triangulation. 

\begin{definition}\label{def:dela_empty_criteria}
  A Delaunay triangulation of a set of $m$ points $\mathbf{P}=\{\mathbf{p}_1,\mathbf{p}_2,...,\mathbf{p}_m\}$ in $\real^{d}$, $\text{DT}(\mathbf{P})$, is any triangulation of $\mathbf{P}$ such that for every $d$-simplex
in $\text{DT}(\mathbf{P})$, the circumscribing hypersphere of the $d$-simplex does not contain any other point of $\mathbf{P}$. 
\end{definition}
Given a set of points in $\real^{d}$, the existence of a unique Delaunay triangulation is based on the following geometric condition: the affine span of $\mathbf{P}$ is $d$-dimensional and no $d+2$ points of $\mathbf{P}$ lie on the same sphere. We refer to such points as points in a \emph{general position}. 

\begin{figure}[ht]
\begin{center}
    \includegraphics[scale=0.6]{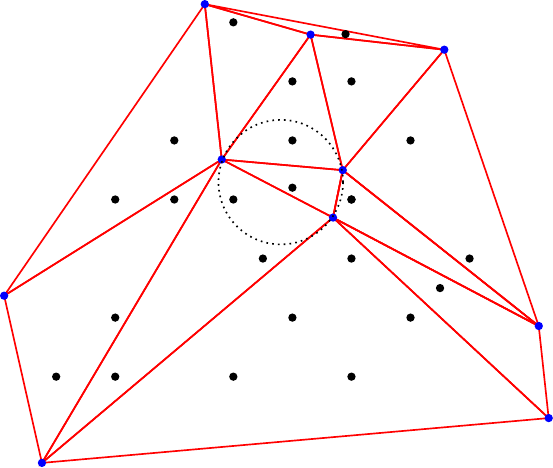}
    \caption{The blue dots indicate the atoms which generate the data points. Each black dot, denoting a data point, is a convex combination of
    three atoms which are vertices of the triangle the point belongs to. Note that the circumscribing circle of any triangle does not contain any additional landmark points.  }
    \label{fig:illustration}
    \end{center}
\end{figure}

\paragraph{Model for generating atoms and data} We consider $m$ landmark points $\a_1,\a_2,...,\a_m$ in $\real^{d}$ with $m\ge d+1$ in general position  meaning that there is a unique Delaunay triangulation. Each data point is in the convex hull of the $m$ landmark points. Figure \ref{fig:illustration} illustrates the model when $d=2$. 

\paragraph{Sparse coding under the Delaunay triangulation model} Let $\A \in \real^{d\times m}$ be the dictionary of the $m$ landmarks defined as $\A=[\a_1,\a_2,...,\a_m]$. Any point $\y$ in the convex hull of the $m$ landmarks can be written as $\y = \A\x$ where $\x \in \Delta^m$. However, note that there may be multiple ways to represent the point $\y$ as a convex combination of the landmark points. Since our aim is to obtain sparse representations, we focus on the problem of finding a unique sparse solution to $\y=\A\x$. Let $\text{DT}(\A)$ denote the set of $d$-simplices that 
constitute the Delaunay triangulation of 
$\{\a_1,\a_2,...,\a_m\}$. For our setting, we define the sparsest representation to be the representation of a point $\y$ using the vertices of the $d$-simplex of $\text{DT}(\A)$ it belongs to. As an example, if $d=2$, this will be representing the point using the vertices of the triangle it belongs to. This motivates the following definition of a weighted $\ell_0$ pseudo-norm.

\begin{definition}
Assume $m$ landmark points $\a_1, \a_2, ..., \a_m$ in $\real^{d}$ have a unique Delaunay triangulation $\text{DT}(\A)$. Let $\y\in \real^{d}$ be an interior point of a $d$-simplex of $\text{DT}(\A)$ with circumcenter $\mathbf{c}$. The \emph{weighted $\ell_0$ norm} of $\x$ is defined as
\begin{equation} \label{eq:weighted_l0_defn}
 \ell_{w,0}(\x) = \frac{1}{\|\x||_0}\sum_{i=1}^{m} \mathbf{1}_{\real_{+}}(x_i) ||\mathbf{c}-\a_i||^2,
 \end{equation}
 where $\mathbf{1}_{\real_{+}}(x_i)=1$ if $x_i>0$ and $0$ otherwise. 
\end{definition}
Given the above definition of  a weighted $\ell_0$ norm and the fact that a given point $\y$ admits different representations as a convex combination of the dictionary atoms, 
the natural question is the sense in which this norm is minimal i.e., among the different representations, which ones admit minimal values in this norm? 
The next theorem shows that the local reconstruction is minimal in the weighted $\ell_0$ norm. The result of Theorem \ref{thm:weighted_lo} follows from the following lemma, which we prove below. 
\begin{lemma}
Let $\A \in \real^{d\times m}$ be the dictionary of the $m$ landmarks defined as $\A=[\a_1,\a_2,...,\a_m]$. Let $\text{DT}(\A)$ denote a set of $d$-simplices of the Delaunay triangulation of $\{\a_1,\a_2,...,\a_m\}$. If $f$ is
a $d$-simplex of $\text{DT}(\A)$ defined by the vertices $\{\a_j : j \in T, |T|=d+1\}$, there is a hypersphere with center $\mathbf{c}$ and radius $R$ such that
$||\a_j-\c||=R$ if $ j\in T$
 and $||\a_j-\c||>R$ if $j\notin T$.
\end{lemma}

\begin{proof}
$f$ is a $d$-simplex of $\text{DT}(\A)$ where the indices of its vertices are in $T$. Let $\c$ and $R$ respectively denote the center and radius of the circumscribing hypersphere of $f$. By construction, $||\a_j-\c||=R$ if $j\in T$. For contradiction, assume that there is a $j \notin T$ such that $||\a_j-\c||\le R$. This contradicts the definition of a Delaunay triangulation in Definition \ref{def:dela_empty_criteria} since $\a_j$ will be an interior point of the circumscribing hypersphere.

\end{proof}

\begin{theorem}\label{thm:weighted_lo}
Given a set of landmarks 
$\{\mathbf{a}_1, \ldots, \mathbf{a}_m\} $ with a unique Delaunay triangulation $\text{DT}(\A)$, let $\y \in \real^{d}$ be an interior point of the $d$-simplex of $\text{DT}(\A)$ with circumcenter $\mathbf{c}$ and radius $R$. 
Let 
\begin{equation*} \label{eq:l0_result}
\x^* = 
 \underset{\x \in \Delta^{m}}{\arg\min}\,\,  \ell_{w,0}(\x) \quad \text{s.t.} \quad \y = \A\x.
\end{equation*}
Then, $\x^*$ is such that $\{j: \x^*_j\neq 0\}$ correspond to the indices of the vertices of the $d$-simplex of $\text{DT}(\A)$ that contains $\mathbf{y}$.
\end{theorem}

\begin{proof}
Consider a $d$-simplex of $\text{DT}(\A)$ containing $\mathbf{y}$ defined by the vertices $\{\a_j : j \in T, |T|=d+1\}$. Using vertices in $T$, $\y$ can be represented as a convex combination using coefficient vector $\x^*$. Let $\x$ be another feasible solution of the program with support $T'$. We now apply Lemma 1 to obtain
\begin{align*}
\frac{1}{||\x||_0} \sum_{i \in T'} \mathbf{1}_{\real_{+}}(x_i) ||\mathbf{c}-\a_i||^2 
&>R^2\sum_{i \in T'} \frac{\mathbf{1}_{\real_{+}}(x_i)}{||\x||_0}\\
&=R^2\\
&= R^2 \sum_{i \in T} \frac{\mathbf{1}_{\real_{+}}(x_i^*)}{||\x^*||_0}\\
&=\frac{1}{||\x^*||_0}\sum_{i \in T}\mathbf{1}_{\real_{+}}(x_i^*)||\mathbf{c}-\a_i||^2
\end{align*}
Therefore, the sparse representation using the vertices in $T$ 
is the optimal solution to the $\ell_{w,0}$ minimization problem. 
\end{proof}

Given a reconstruction $\y = \A\x$, we note that the weighted $\ell_0$ norm puts a uniform prior on all atoms which are used in the representation. However, there are two drawbacks of this regularization. First, the definition of the weighted $\ell_0$ norm depends on knowing the circumcenter of the $d$-simplex the point belongs to. In addition, the regularizer uses an indicator function which is not suitable for optimization. To obtain a regularization amenable to optimization,
we now define
a convex relaxation of the weighted $\ell_0$ problem as follows. 
\begin{definition}
Assume $m$ landmark points $\a_1, \a_2, ..., \a_m$ in $\real^{d}$ with a unique Delaunay triangulation $\text{DT}(\A)$. Let the point $\y\in \real^{d}$ be in the convex hull of the landmark points i.e., $\y = \A\x$ with $\x\in \Delta^m$. 
The \emph{weighted $\ell_1$} norm of $\x$ is defined as
\begin{equation} \label{eq:weighted_l1_defn}
 \ell_{w,1}(\x) = \sum_{i=1}^{m} x_i\, ||\y-\a_i||^2,
 \end{equation}
\end{definition}

Analogous to compressed sensing theory, the next question is the sense in which a weighted $\ell_1$ minimization is equivalent to a weighted $\ell_0$ minimization problem. This equivalency is summarized in Theorem \ref{thm:weighted_l1_uniqueness}. The following lemma will be essential to the proof of Theorem \ref{thm:weighted_l1_uniqueness}.
    
\begin{lemma}
Given the dictionary of landmarks $\A \in \real^{d\times m}$, let $\y = \A\x$ for $\x \in \Delta^m$. For any arbitrary point $\c \in \real^{d}$, 
\[
\ell_{w,1}(\x) = \sum_{i=1}^{m} x_i\, ||\y-\a_i||^2 = \sum_{i=1}^{m} x_i ||\a_i-\c||^2 -||\y-\c||^2. 
\]

\end{lemma}

\begin{proof}
    We expand $\ell_{w,1}(\x)$ as follows and use the fact that $\y= \A\x$ and $\x \in \Delta^m$:
    \begin{align*}
& \sum_{i=1}^{m} x_i\, ||\y-\a_i||^2\\
& = \sum_{i=1}^{m} x_i\, ||(\y-\c)+(\c-\a_i)||^2\\
& = \sum_{i=1}^{m} x_i\, \left(||\y-\c||^2 +||\a_i-\c||^2 - 2(\y-\c)^T(\a_i-\c)\right)\\
& = ||\y-\c||^2 + \sum_{i=1}^{m} x_i||\a_i-\c||^2 -2(\y-\c)^T\left(\sum_{i=1}^{m}x_i\a_i-\c\right)\\
& = ||\y-\c||^2 + \sum_{i=1}^{m} x_i||\a_i-\c||^2 -2||\y-\c||_2^2\\
& = \sum_{i=1}^{m} x_i ||\a_i-\c||^2 -||\y-\c||^2.
    \end{align*}

\end{proof}

\begin{theorem}\label{thm:weighted_l1_uniqueness}
Given a set of landmarks 
$\{\mathbf{a}_1, \ldots, \mathbf{a}_m\} $ with a unique Delaunay triangulation $\text{DT}(\A)$, let $\y \in \real^{d}$ be an interior point of a $d$-simplex of $\text{DT}(\A)$. Let
\begin{equation*} \label{eq:main_opt}
\x^* = 
\underset{\mathbf{x} \in \Delta^{m}}{\arg\min}\,\,
\sum_{i} x_{i}\|\mathbf{y} - \mathbf{a}_i\|^2 \quad
\text{s.t.}\quad  \mathbf{y} = \A\x.
\end{equation*}
Then, $\x^{*}$ is such that $\{i: \x^*_i\neq 0\}$ correspond to the indices of the vertices of the $d$-simplex of $\text{DT}(\A)$ that contains $\mathbf{y}$.

\end{theorem}

\begin{proof}
Consider the $d$-simplex containing $\mathbf{y}$ defined by the vertices $\{\a_j : j \in T, |T|= d+1\}$. Since $\y$ is an interior point of the $d$-simplex, it can be represented as a convex combination of its vertices using coefficient vector $\x^*$. Note that $\x^*$ is supported on $T$ with $||\x^*||_0= d+1$. Let $\x$ be another feasible solution of the program with support $T'$. We now apply Lemma 3 to $\y = \A\x$ with $\c$ as the circumcenter of the $d$-simplex that contains $\y$: 
\[
 \sum_{j \in T'} x_j \|\mathbf{y} - \mathbf{a}_j\|^2 = \sum_{j \in T'} x_j \|\mathbf{a}_j - \mathbf{c}\|^2 - \|\mathbf{y} - \mathbf{c}\|^2. 
\]
We now apply Lemma 1 to lower bound the above term. Specifically, we use the fact that $||\a_j-\c||>R$ if $j\notin T$ and $||\a_j-\c||=R$ if $j\in T$:
\begin{align*}
\sum_{j \in T'} x_j \|\mathbf{y} - \mathbf{a}_j\|^2 &= \sum_{j \in T'} x_j \|\mathbf{a}_j - \mathbf{c}\|^2 - \|\mathbf{y} - \mathbf{c}\|^2. \\
&> \sum_{j \in T'} x_j R^2 - \|\mathbf{y} - \mathbf{c}\|^2. \\
&= R^2 - \|\mathbf{y} - \mathbf{c}\|^2. \\
&= \sum_{j \in T} x^*_j ||\a_j-\c||^2- \|\mathbf{y} - \mathbf{c}\|^2. \\
& = \sum_{j \in T} x^*_j \|\mathbf{y} - \mathbf{a}_j\|^2.
\end{align*}
Above, the inequality in the second line uses the fact that there is at least one index in $T'$ that is not in $T$. The last equality follows from applying Lemma 3 with $\y = \A\x^*$ and $\c$ as the circumcenter. We have established that $\ell_{w,1}(\x)>\ell_{w,1}(\x^*)$ for any feasible $\x$. Therefore, the sparse representation using the vertices in $T$ 
is the optimal solution to the $\ell_{w,1}$ minimization problem.  

\end{proof}

\subsection{Stability analysis}

In this section, we consider the stability of sparse representations when an input data is perturbed by a bounded additive noise. Formally, given a data point $\y \in \real^{d}$, the data is perturbed resulting $\tilde{\y}$ with the condition that $||\y-\tilde{\y}||_2\le \epsilon$. In the analysis to follow, the notion of a local dictionary is used which we define below.

\begin{definition}
Given a set of landmarks 
$\{\mathbf{a}_1, \ldots, \mathbf{a}_m\} $ with a unique Delaunay triangulation $\text{DT}(\A)$, let $\y\in \real^{d}$ be interior points of the $d$-simplex of $\text{DT}(\A)$. 
Then the local dictionary $\A_{L}\in \real^{d\times d+1}$ associated to $\y$ is
$
\A_{L} = [\a_{j_1}\,\a_{j_2}\,...\a_{j_{d+1}}],
$
where the indices $\{j_k\}_{k=1}^{d+1}$ correspond to the vertices of the d-simplex that contains $\y$. 
\end{definition}
The utility of a local dictionary is that it allows us to express a data point in terms of its barycentric coordinates.
\begin{definition}
    Given a local dictionary $\A_L \in \real^{d\times (d+1)}$ associated to $\y\in \real^{d}$, the barycentric coordinates of $\y$ is the unique solution to the linear system $\B_{L}\x = \z$,
where $\B_L\in \real^{(d+1)\times (d+1)}$ is defined as $\B_L= \begin{pmatrix}\A_L\\ \mathbf{1}_{d} \end{pmatrix}$ and $\z = \begin{pmatrix}\y \\ 1\end{pmatrix}$.
\end{definition}

\begin{theorem}\label{thm:weighted_l1_uniqueness_stability}
Given a set of landmarks 
$\{\mathbf{a}_1, \ldots, \mathbf{a}_m\} $ with a unique Delaunay triangulation $\text{DT}(\A)$, let $\y,\tilde{\y} \in \real^{d}$ be interior points of the same $d$-simplex of $\text{DT}(\A)$. Further, assume that $||\y-\tilde{\y}||\le \epsilon$ and $\y = \A\x^*$ where 
\begin{equation*} 
\x^* = 
\underset{\mathbf{x} \in \Delta^{m}}{\arg\min}\,\,
\sum_{j} x_{j}\|\y - \mathbf{a}_j\|^2 \quad
\text{s.t.}\quad  \y = \A\x.
\end{equation*}
Let $\tilde{\x}^*$ be the optimal solution to the following $ \ell_{w,1}$ minimization problem.
\begin{equation*} \label{eq:main_opt_stability}
\tilde{\x}^* = 
\underset{\mathbf{x} \in \Delta^{m}}{\arg\min}\,\,
\sum_{j} x_{j}\|\tilde{\y} - \mathbf{a}_j\|^2 \quad
\text{s.t.}\quad  \tilde{\y} = \A\x.
\end{equation*}
Then, $\tilde{\x}^*$ to the above program is such that
\[
||\tilde{\x}^*-\x^*|| \le\frac{1}{\sigma_{\min}(\B_L)}\,\epsilon,
\]
where $\B_L= \begin{pmatrix}\A_L\\ \mathbf{1}_{d} \end{pmatrix}$ and $\A_L$ is the local dictionary associated to $\tilde{\y}$.
\end{theorem}

\begin{proof}
Since $\y$ and $\tilde{\y}$ belong to the same simplex of $\text{DT}(\A)$, they have the same local dictionary denoted by $\A_L$. Using Theorem 4, the optimal solution $\tilde{\x}^*$ is such that it is only nonzero on the indices corresponding to vertices of the simplex that contains $\tilde{\y}$. An analogous argument could be made for $\x^*$. It then follows that $\y= \A_L\x^*$ and $\tilde{\y}=\A_{L}\tilde{\x}^*$. In what follows, we form a square linear system by considering an additional constraint that the coefficients must sum to $1$. To that end, we define $\z,\tilde{\z}\in \real^{d+1}$ as follows: $\z = \begin{pmatrix} \y\\1 \end{pmatrix}$ and $\tilde{\z} = \begin{pmatrix} \tilde{\y}\\1 \end{pmatrix}$. Note that $||\z-\tilde{\z}||_2=||\y-\tilde{\y}||_2$. Further, $\z = \B_L\x^*$ and $\tilde{\z}=\B_{L}\tilde{\x}^{*}$. We proceed to lower bound  $||\z-\tilde{\z}||_2$:
\[
||\z-\tilde{\z}||_2 = ||\B_{L}(\x^*-\tilde{\x}^{*})||_2\ge \sigma_{\min}(\B_L)||\x^*-\tilde{\x}||_2,
\]
where $\sigma_{\min}(\B_L)>0$ (this follows from the assumption that the landmarks are in general position). Combining this lower bound with $||\y-\tilde{\y}||_2\le \epsilon$, we obtain
\[
||\tilde{\x}^*-\x^*|| \le\frac{1}{\sigma_{\min}(\B_L)}\,\epsilon.
\]

    \end{proof}

\noindent \textbf{Remark}: We would like to highlight that the affine constraint on the coefficients ensures that the aforementioned theorem remains valid even when the data points are translated. However, the stability of the theorem is contingent upon the minimum singular value of a shifted $\B_L$. We note that when the noise is sufficiently low and $\sigma_{\min}(\B_L)$ is appropriately large, the stability analysis ensures a robust sparse solution. This robustness depends upon the magnitude of $\sigma_{\min}(\B_L)$, which in turn is influenced by the geometrical structure of the localized dictionary. Initial numerical experiments suggest that if the localized dictionaries are ``well-structured", $\sigma_{\min}(\B_L)$ tends to be relatively large, whereas smaller values of $\sigma_{\min}(\B_L)$ correspond to elongated triangles. Details on the numerical experiments can be found in the Supplementary Materials.

\subsection{Optimal dictionary}
In the theoretical analysis so far, we have studied the problem of recovering 
a sparse coefficient vector given a fixed dictionary. In this section, we assume that the sparse coefficients are fixed and study the optimization problem over the dictionary. In particular, we study the optimal solution defined as follows. 
\begin{equation} \label{eq:min_wrt_A}
\A^* = 
\underset{\A \in \real^{d\times m}}{\arg\min}\,\, ||\Y-\A\X||_F^2 + \lambda \sum_{i=1}^{n}\sum_{j=1}^{m} (\x_i)_j ||\y_i-\a_j||^2,
\end{equation}
where $\lambda>0$ is a regularization parameter. Below, we will prove that $\A^*$ is unique and has a closed form solution.  
\begin{theorem}\label{thm:opt_over_atoms}
For fixed $\X \in S$, $\A^*$ is given by
\[
\A^* = (1+\lambda) \Y\X^T\mathbf{H}^{-1},
\]
where $\mathbf{H} = \X\X^T+\lambda \text{diag}(\X\mathbf{1})$.

\end{theorem}

\begin{proof}

Let $f(\A)$ denote the objective function in \eqref{eq:min_wrt_A}. The proof of the theorem relies on showing that $f(\A)$ is strongly convex. Some calculation yields $\nabla f(\A) = 2(\A\X-\Y)\X^T + 2\lambda \A \text{diag}(\X\mathbf{1})-2\lambda \Y\X^T$. Strong convexity requires showing that $ \langle \nabla f(\A_1)-\nabla f(\A_2),\A_1-\A_2\rangle \ge \mu ||\A_1-\A_2||_F^2$ with $\mu>0$ for any $\A_1,\A_2$. Using the explicit form of the gradient, strong convexity is equivalent to showing that $\langle \X\X^T+\lambda\text{diag}(\X\mathbf{1}), (\A_1-\A_2)^T(\A_1-\A_2)\rangle \ge \mu ||\A_1-\A_2||_F^2$. For ease of notation, let $\mathbf{H} = \X\X^T+\lambda\text{diag}(\X\mathbf{1})  $ and $\mathbf{G} = (\A_1-\A_2)^T(\A_1-\A_2)$. We first note that $\mathbf{H}$ is symmetric positive definite and $\mathbf{G}$ is symmetric positive semidefinite. To see the former claim, it suffices to show that the diagonal entries of $\text{diag}(\X\mathbf{1})$ are non-zero. The only case an entry will be zero is if an atom is not used by all data points. For this case, the given atom can be discarded. In all other cases, all the diagonal entries of $\text{diag}(\X\mathbf{1})$ are positive. Finally, we claim that $\langle \mathbf{H}\,,\mathbf{G}\rangle \ge \lambda_{\min}(\mathbf{H})\,\text{trace}(\mathbf{G})$. This gives the desired strong convexity result with $\mu = \lambda_{\min}(\mathbf{H})>0$. Setting the gradient to zero yields the unique solution $\A^{*} = (1+\lambda) \Y\X^T\mathbf{H}^{-1}$. It remains to prove the claim that $\langle \mathbf{H}\,,\mathbf{G}\rangle\ge \lambda_{\min}(\mathbf{H})\,\text{trace}(\mathbf{G})$. This follows from noting that the matrix $\mathbf{H}-\lambda_{\min}(
\mathbf{H})\mathbf{I}$ is symmetric positive semidefinite and the term $\langle \mathbf{H}-\lambda_{\min}(\mathbf{H})\mathbf{I}\,,\mathbf{G}\rangle \ge 0$ as it is a trace product of symmetric positive semidefinite matrices.\\* 

\end{proof}

\noindent \textbf{Remarks}: We note that the weighted $\ell_1$ regularizer enables us to obtain strong convexity when optimizing over the dictionary atoms. If $\lambda=0$, strong convexity is not always guaranteed. We also note that each column of the optimal dictionary is a linear combination of the data points. 

\subsection{KDS embedding}

In a typical setting, under the manifold hypothesis, the number of landmarks is expected to be much smaller than the number of data points i.e., $m\ll n$.  With that, the optimal sparse coefficients obtained from solving \eqref{eq:optimization_alg} are a low-dimensional representation of the high-dimensional data. However when utilizing the sparse coefficients for downstream tasks such as clustering, further dimensionality reduction can be useful. For instance, this will be the case in the setting where $m\ll n$, such that the data is well represented via local landmarks, but $m\gg k$ (e.g., $k$ is number of clusters). In what follows, using connections to spectral clustering and spectral embedding \citep{ng2001spectral,belkin2003laplacian}, we will show how to obtain low-dimensional embeddings based on the eigenvectors of the covariance matrix $\X\X^T$. 

The starting point is the observation that the representation coefficients $\X$ define a bipartite \emph{similarity graph} $G$ with $n+m$ vertices corresponding to the $n$ data points and $m$ learned dictionary atoms. In this graph, each data point $\mb{y}_i$ and each atom $\mb{a}_j$ is connected by an undirected edge of weight $(\x_{i})_j$. To embed the data points and the atoms into $\real^{k}$, we consider the classic spectral embedding.
\begin{equation}
    \underset{\Q \in \real^{k\times (n+m)}}{\min}\,\, \text{trace}(\Q\mathbf{L}\Q^T)\quad \text{s.t.}\,\, \Q\Q^T=\I,
\end{equation}
where $\mb{Q} = [\mb{Q}_{\mb{Y}} \,\,\mb{Q}_{\mb{A}}] \in \real^{k \times (n + m)}$. We enforce an additional constraint $\mb{Q}_{\mb{Y}} = \mb{Q}_{\mb{A}}\mb{X}$ to formulate the problem only in terms of the landmarks.
We note that this type of assumption has been used for landmark-based locally linear embedding \citep{vladymyrov2013locally}. We will now proceed to state and prove a lemma which shows that the Laplacian quadratic form could be formulated in terms of the landmarks. The Schur complement will be used in the proof and is defined as follows. Consider the block matrix $\mathbf{M}=\begin{pmatrix}
    \A& \B\\
    \C&\D
\end{pmatrix}$
where $\A \in \real^{p\times p}$,
$\B\in \real^{p \times q}$, $\C\in \real^{q \times p}$ and $\D\in \real^{q\times q}$. 
If $\A$ is invertible, the Schur complement of $\mathbf{M}$ with respect to $\A$ is defined as $\D-\C\A^{-1}\B$. 
\begin{lemma}
Let $\mb{Q} = [\mb{Q}_{\mb{Y}} \,\,\mb{Q}_{\mb{A}}] \in \real^{k \times (n + m)}$. If $\Q_{\Y}= \Q_{\A}\X$,
\[
\mathrm{trace}(\Q\mathbf{L}\Q^T) =  \mathrm{trace}\left(\Q_{\A}\mathbf{L}_{\A}
\Q_{\A}^T\right).
\]
\end{lemma}
\begin{proof}
Using the weight matrix in \eqref{eq:laplacian_weight}, the Laplacian $\L$ is given by $\begin{bmatrix}
  \mathbf{I}
  & \rvline & -\X^T \\
\hline
  -\X & \rvline &
  \text{diag}(\X\mathbf{1})
\end{bmatrix}$. We now proceed to evaluate $\mathrm{trace}(\Q\mathbf{L}\Q^T)$. 
\begin{align*}
&\text{trace}(\Q\mathbf{L}\Q^T) =
 \text{trace}(\Q^T\Q\mathbf{L})\\
=&\,\text{trace}\left(\begin{bmatrix}
  \Q_{\Y}^T\Q_{\Y}\vspace{0.1em}
  & \rvline & \Q_{\Y}^T\Q_{\A}\vspace{0.1em} \\
\hline
  \vspace{0.2em}\Q_{\A}^T\Q_{\Y} & \rvline &
  \vspace{0.2em} \Q_{\A}^T\Q_{\A}
\end{bmatrix}
\begin{bmatrix}
  \mathbf{I}
  & \rvline & -\X^T \\
\hline
  -\X & \rvline &
  \text{diag}(\X\mathbf{1})
\end{bmatrix}\right)\\
 = &\text{trace}\left((\Q_{\Y}^T\Q_{\Y})\I-\Q_{\Y}^T\Q_{\A}\X-\Q_{\A}^T\Q_{\Y}\X^T+\Q_{\A}^T\Q_{\A}\mathbf{J}\right)\\
  = &\text{trace}\left(\X^T\Q_{\A}^T\Q_{\A}\X-2\X^T\Q_{\A}^T\Q_{\A} \X+\Q_{\A}^T\Q_{\A}\mathbf{J}\right)\\
  = &\text{trace}\left(\Q_{\A}^T\Q_{\A}
  \left(\mathbf{J}-\X\X^T\right)\right)
 =  \text{trace}\left(\Q_{\A}\mathbf{L}_{\A}\Q_{\A}^T\right),
\end{align*}
where $\mathbf{J} = \text{diag}(\X\mathbf{1})$ and $\mb{L}_{\mb{A}}$ is known as the \emph{Schur complement} of $\mb{L}$ with respect to $\mb{Y}$. 
\end{proof}

Given the above proof, we consider the following spectral embedding problem
\begin{equation}
    \underset{\Q_{\A} \in \real^{k\times m}}{\min}\,\, \text{trace}(\Q_{\A}\mathbf{L}_{\A}\Q_{\A}^T)\quad \text{s.t.}\,\, \Q_{\A}\Q_{\A}^T=\I.
\end{equation}
The above problem is a standard spectral problem whose optimal solution is $\Q_{\A}^{*} = \mathbf{U}_k^T$ where the columns of $\mathbf{U}_k\in \real^{m\times k}$ are the eigenvectors of $\mathbf{L}_{\A}$ corresponding to the largest $k$ eigenvalues. It follows that the dominant computation of the KDS spectral embedding only requires the calculation of the first $k$ eigenvectors of an $m \times m$ matrix $\mathbf{L}_{\A}$, which is very small when $m \ll n$, as well as a handful of $O(mn)$-time multiplications by the matrix $\mb{X}$ to compute the adjacency matrix $\mb{X}\mb{X}^\top$ and recover $\mb{Q}_{\mb{Y}} = \mb{Q}_{\mb{A}}\mb{X}$. 

We note that it is important to set $k$ and $m$ carefully. In lack of prior knowledge about number of clusters, one could employ the eigengap heuristic \citep{von2007tutorial} which sets number of clusters based on the gap between eigenvalues of the graph Laplacian. In terms of $m$, a relatively large value of $m$, implies that points are well represented via local landmarks. However, this has the implication that points within the same cluster may not have the same sparsity structure. In contrast, a relatively small value of $m$ would allow points from different clusters to have a similar sparsity structure (which leads to sub-optimal clustering). 

\section{Dictionary learning algorithm}\
\label{sec:opt}

In this section, given a set of data points, we discuss the problem of estimating both the sparse representations and dictionary atoms. To this end, we study the following minimization problem:
\begin{equation}\label{eq:main_kds_opt}
\underset{\A \in \real^{d\times m},\X \in S}{\min}\,\,\,  \sum_{i=1}^{n} \bigg[  \frac{1}{2}\|\y_i - \A \x_i \|^2 + \lambda \sum_{j=1}^m (\x_{i})_j\|\y_i - \a_j\|^2\bigg],
\end{equation}  
where $\X =[\x_1,\x_2,...,\x_n]$ with each $\x_i \in \Delta^{m}$. The balance between the reconstruction loss and the locality regularization is controlled by the parameter $\lambda$. A standard way to solve the above minimization program is alternating minimization which alternates between sparse approximation and dictionary update steps \citep{agarwal2016altmin}. We discuss the two steps below. The KDS algorithm is summarized in in Algorithm \ref{alg:alt_min}.

\subsection{Sparse coding} 

Given a fixed dictionary $\A$, the sub-problem over the sparse coefficients is a weighted $\ell_1$ minimization problem for which efficient methods exist \citep{asif2012weightedl1}. We consider
the accelerated projected gradient descent algorithm \citep{su2014differential} to solve this problem. Since the minimization problem for $\X$ decouples into optimizing the sparse representation of each data point, we consider the problem of finding the optimal coefficient given the dictionary $\A$ and a data point $\y$ as follows
\begin{equation}
\label{eq:sparse_coding}
\x^*(\A, \y) =  \argmin_{\x \in \Delta^{m}}  \,\, \frac{1}{2}\|\y - \A \x \|^2 + \lambda \sum_{j=1}^m x_{j}\|\y - \a_j\|^2.
\end{equation}
Let $\L(\A,\y,\x,\lambda)$ denote the objective in the above program. \\*

\noindent \textbf{The accelerated projected gradient descent}: This method starts with the initialization $\x^0=\tilde{\x}^{(0)} = \mathbf{0}$ and considers the following updates
\begin{align*}
\x^{(t+1)} &= \mathcal{P}_{\Delta^{m}}\left(\tilde{\x}^{(t)} - \alpha \nabla_{\x} \L(\A, \y, \tilde{\x}^{(t)})\right) \\
\tilde{\x}^{(t+1)} &= \x^{(t + 1)} + \frac{t-1}{t+2}(\x^{(t + 1)} - \x^{(t)}).
\end{align*}
for $0 \le t \le T_{\max}$. The operator $\mathcal{P}_{\Delta^{m}}$ projects onto $S$, the probability simplex and has a closed form that can be readily computed \citep{wang2013projection,condat2016fast}. The parameter $\alpha$ is a step size. We note below the gradient of $\L$ with respect to $\x_i$:
\begin{equation*}
\nabla_{\x_i} \L(\A, \y_i, \x_i,\lambda) = \A^\top(\A\x_i - \y_i) + \lambda \sum_{j=1}^m \|\y_i - \a_j\|^2 \e_j
\end{equation*}

\subsection{Dictionary learning} 

After $T_{\max}$ iterations of the sparse coding step, we have optimized sparse coefficients $\{\x_i^{(T_{\max})}\}_{i=1}^{n}$ corresponding to the data points  $\{\y_i\}_{i=1}^{n}$. The next part of the algorithm is to optimize for the dictionary which can be estimated by solving the following optimization problem:
\begin{equation}\label{eq:main_opt_relax}
\underset{\A \in \real^{d\times m}}{\min}\,\,\,  \sum_{i=1}^{n} \bigg[  \frac{1}{2}\|\y_i - \A \x_i^{(T_{\max})} \|^2 + \lambda \sum_{j=1}^m (\x_i^{(T_{\max})})_{j}\|\y_i - \a_j\|^2\bigg].
\end{equation}
Let $\L_1(\A,\y,\x,\lambda)$ denote the objective in the above program.
We note that the gradient of $\L_1$ with respect to $\A$ is given by
\[
\nabla_{\A} \L_1 = 2(\A\X-\Y)\X^T + 2\lambda \A \text{diag}(\X\mathbf{1})-2\lambda \Y\X^T
\]
The dictionary learning sub-problem can be solved using gradient descent. 

\subsection{Complexity of alternating minimization}
For a fixed data point, the gradient update to estimate the coefficient is $O(md)$ and the projection onto the simplex is $O(m\log(m))$ \citep{wang2013projection}.
Therefore, the per-iteration cost of sparse coding is $O(nm\max(\log(m),d))$. The per-iteration complexity of the dictionary learning step is $O(nmd)$ which
is the cost of the gradient update. 

\begin{algorithm}
\caption{KDS algorithm to solve \eqref{eq:main_kds_opt}}
\label{alg:alt_min}
\begin{algorithmic}[1]
\State \textbf{Input:} Data points $\Y = [\y_1, \ldots, \y_n] \in \real^{d \times n}$, maxiterations.
\State \textbf{Initialization:} $\x_i^{(0)} = \mathbf{0}$ for $1\le i\le n $. Set $\A^{(0)}\in \real^{d\times m}$ to be random subset of data. 
\For {k = 1:maxiterations}
\State Set step size: $\alpha = \frac{1}{(\sigma_{\max}(\A^{(k-1)}))^2}$.  
\State \textbf{Sparse coding via encoder}: Given $\A^{(k-1)}$, use accelerated project gradient descent to obtain $\{\x_1^{(k)},\x_2^{(k)},...,\x_n^{(k)}\}$.
\State \textbf{Decoder}: Reconstruct approximate data $\{\A^{(k-1)}\x_1^{(k)},...,\A^{(k-1)}\x_n^{(k)}\}$.
\State \textbf{Dictionary learning}: Backpropagation to obtain $\A^{(k)}$.    \EndFor
\end{algorithmic}
\end{algorithm}

\subsection{Algorithm unrolling} 
In order to solve \eqref{eq:main_kds_opt} efficiently and to design an interpretable network, we consider a technique known as algorithm unrolling. This is the process of designing a highly-structured recurrent neural network to efficiently solve problems \citep{Monga2021}. Although our application of the technique for manifold learning is new, there exists a rich literature on the subject in the context of sparse dictionary learning \citep{chang2019randnet,gregor2010learning,rolfe2013discriminative,tolooshams2018scalable,tolooshams2020deep,tolooshams2020convolutional,tolooshams2022stable}. In order to solve the relaxed optimization problem in \eqref{eq:main_opt_relax}, we introduce an autoencoder architecture that implicitly solves the problem when trained by backpropagation. Given a dictionary $\A$, our encoder maps a data point $\y$, or a batch of such points, to the sparse code $\x$ minimizing $\L(\A, \y, \x)$. This is done by unfolding $T$ iterations of projected gradient descent on $\L$ into a deep recurrent neural network. Our linear decoder reconstructs the input as $\hat{\y} = \A \x$. The network weights correspond to the dictionary $\A$, which is initialized to a random subset of the data $\Y$ and then trained to minimize \eqref{eq:main_kds_opt} by backpropagation through the entire autoencoder.  If we view the forward pass through our encoder  as an analogue of the sparse recovery step used in traditional alternating-minimization schemes, then this backward pass corresponds to an enhanced version of the so-called ``dictionary update'' step. We note that the projection onto the probability simplex can be written as a modified ReLU function and thus serves as a non-linear activation function in the encoder.

\section{Experiments}
\label{sec:experiments}

\subsection{Application of KDS to clustering}
Let $\Y = [\y_1, \ldots, \y_n] \in \real^{d \times n}$ be a collection of $n$ data points in $\real^d$. To cluster the data, we utilize Algorithm 1 to obtain sparse representation coefficients $\X$ and a set of $m$ atoms $\mb{a}_1, \ldots, \mb{a}_m$. Our similarity matrix is $\X\X^T$.
Given the similarity matrix, to cluster the data into $k$ clusters, we apply spectral clustering which first embeds the data using $k$ eigenvectors of a normalized graph Laplacian corresponding to the largest $k$ eigenvalues. We note that the obtained embedding is extended to all data points by applying the dictionary. The details of these are in Section IV. D. Given the embedding, we run $k$-means to obtain the cluster labels \citep{ng2001spectral}. 

In this section, we demonstrate the ability of KDS, implemented in PyTorch \citep{paszke2019pytorch}, to efficiently and accurately recover the underlying clusters of both synthetic and real-world data sets.  Details about pre-processing of data and parameter selection for KDS as well as baseline algorithms can be found in the Supplementary Materials. All clustering experiments are evaluated with respect to a given ground truth clustering using the unsupervised clustering accuracy (ACC), which is invariant under a permutation of the cluster labels. Accuracy is defined as the percentage of correct matches
with respect to the ground truth labels of the data.

\subsection{Synthetic Data}

\begin{figure}
\centering
    \includegraphics[width=0.24\linewidth]{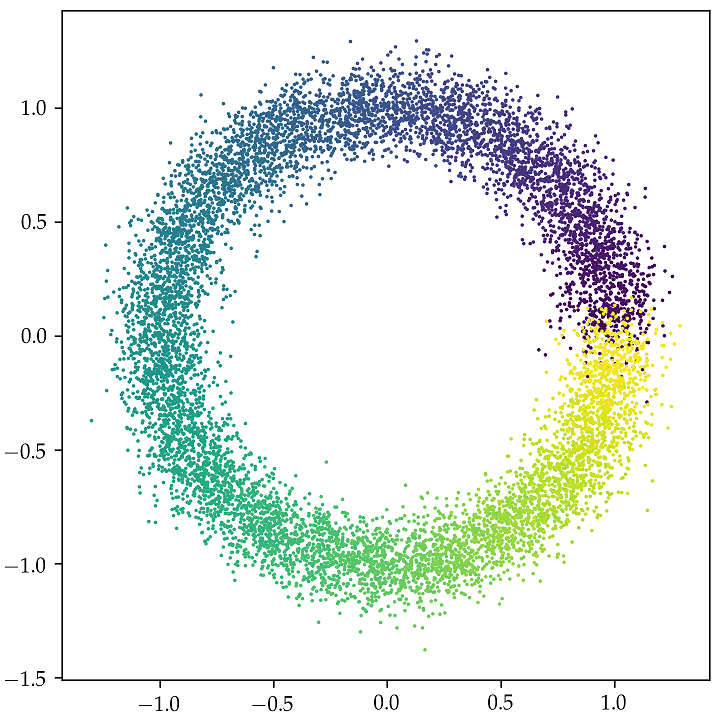}
    \includegraphics[width=0.24\linewidth]{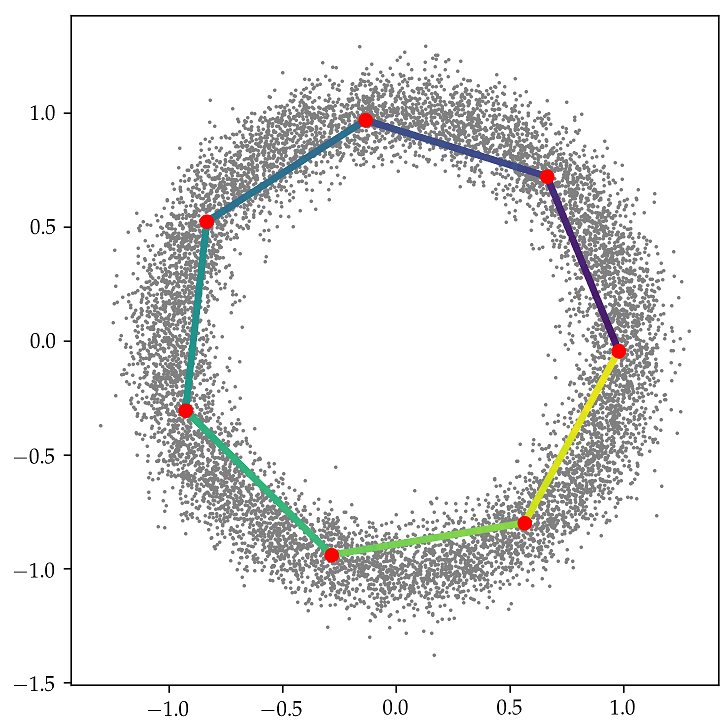} 
    \includegraphics[width=0.24\linewidth]{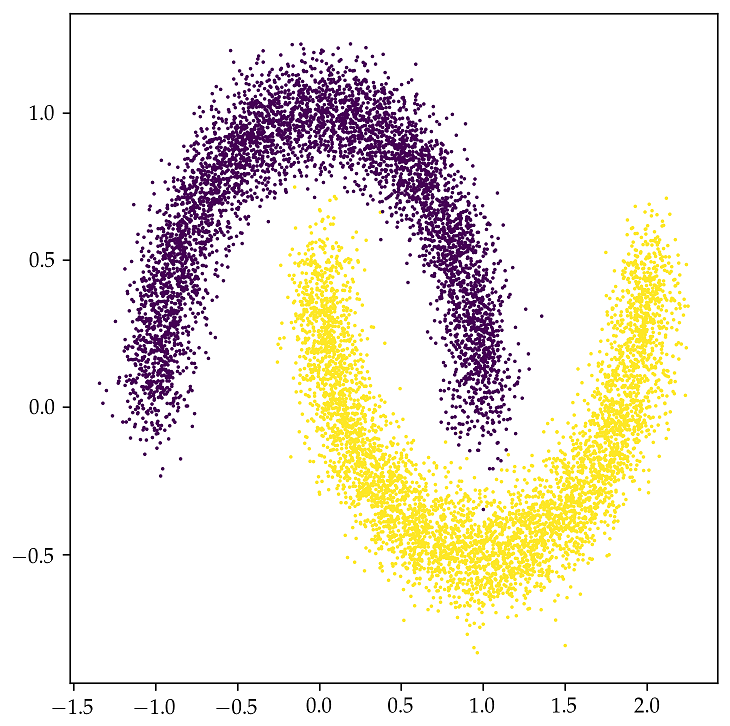}
    \includegraphics[width=0.24\linewidth]{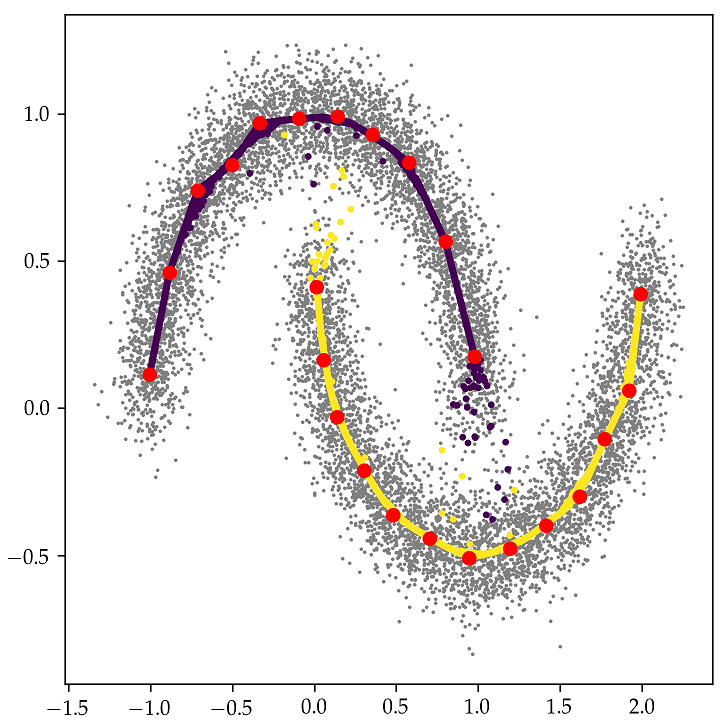}
    \caption{Circle and two moons. Autoencoder input (first and third) and output (second and fourth), with learned atoms marked in red.}
    \label{fig:synthetic}
\end{figure}

\textbf{Learned Dictionary Atoms:}  For our first experiment, we visualize the dictionary atoms learned by our autoencoder when the data is sampled from one-dimensional manifolds in $\real^2$. Figure \ref{fig:synthetic} shows two such data sets. 

The first is the unit circle in $\real^2$. The second is the classic two moon data set \citep{ng2001spectral}, which consists of two disjoint semicircular arcs in $\real^2$. For each of these two data sets, we trained the autoencoder on data sampled uniformly from the underlying manifold(s). We added small Gaussian white noise to each data point to make the representation learning problem more challenging. Figure \ref{fig:synthetic} shows the result of training the autoencoder on these data sets. We see that in each case, the atoms learned by the model are meaningful. Moreover, in each case, we accurately reconstruct each data point as sparse convex combinations of these atoms, up to the additive white noise. As a final remark, drawing a sample of $5000$ data points from the noisy two moons distribution, computing their sparse coefficients, and performing spectral clustering with $k=2$ on the associated bipartite similarity graph results in a clustering accuracy of $99.9\%$. We note that KDS outperforms baseline algorithms (see Table \ref{tab:acc}). 

\textbf{Clustering with Narrow Separation:}  Our next experiment assesses the clustering capabilities of our algorithm in a toy setting. We studied a simple family of data distributions consisting of two underlying clusters in $\real^2$. These clusters took the shape of two concentric circles of radii $r_{\mr{outer}} = 1$ and $r_{\mr{inner}} = 1 - \delta$, where $\delta \in [0, 1]$ is a separation parameter. For multiple values of $\delta$, we trained our structured autoencoder with $m$ atoms on data sampled uniformly from these two manifolds, each with half the probability mass. For this experiment, we did not add any Gaussian noise to the data.

\begin{figure}[htbp!]
    \centering
    \includegraphics[width=0.95\linewidth]{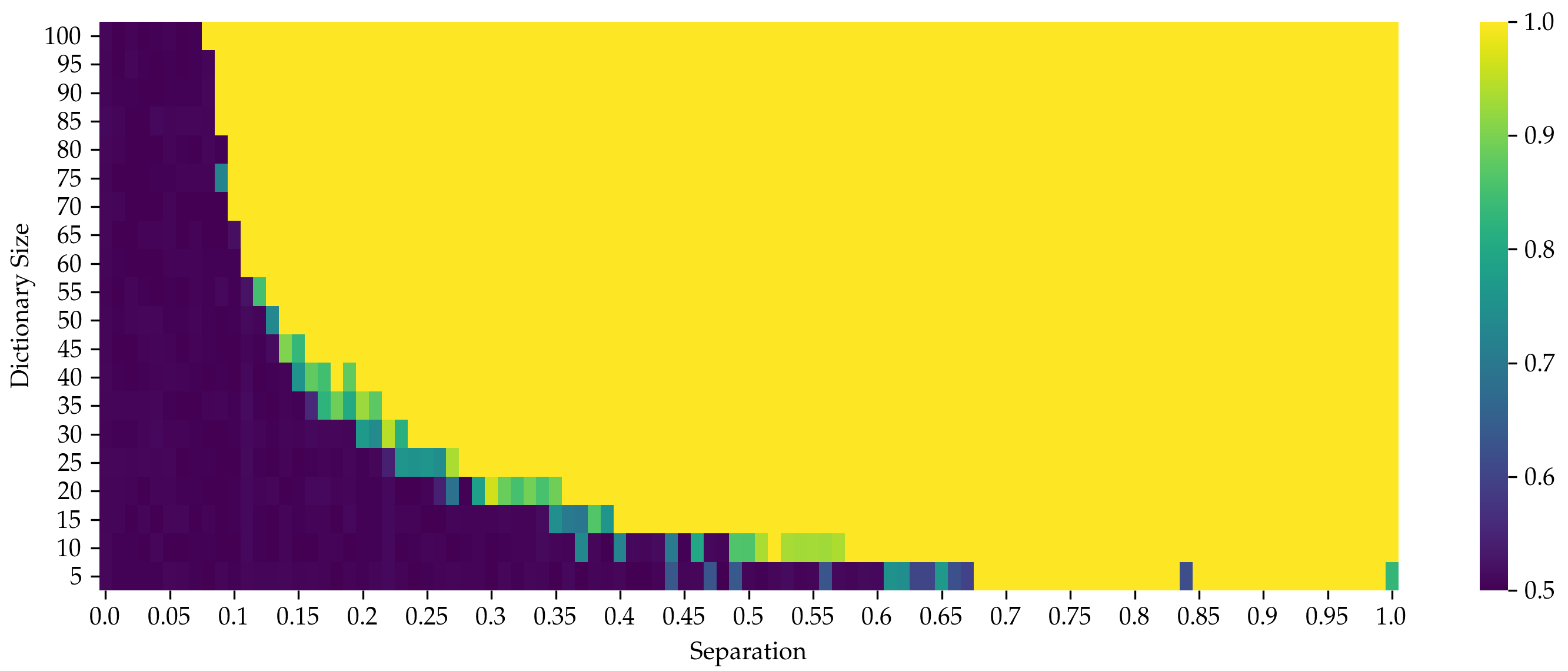}
    \caption{Clustering accuracy for concentric circles across $\delta,m$.}
    \label{fig:concentric_circles_accuracy}
\end{figure}

\begin{figure*}[htbp]
    \centering
    \includegraphics[width=0.9\linewidth]{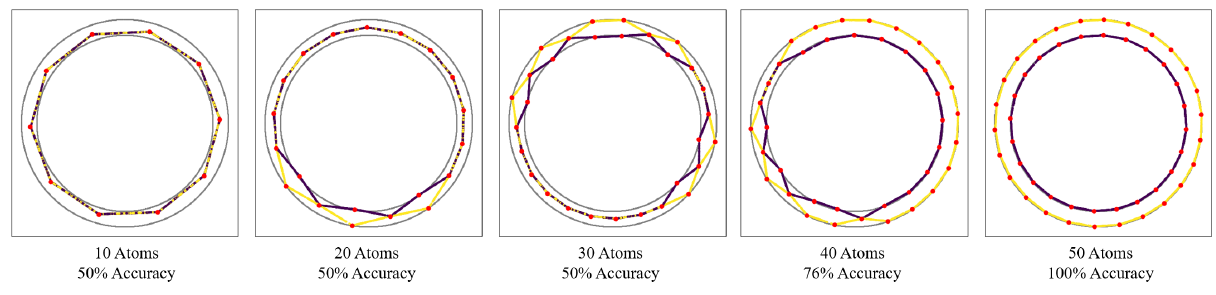}
    \caption{(a) Autoencoder output and learned atoms for concentric circles, separation $\delta = 0.15$.}
    \label{fig:concentric_circles_visualization}
\end{figure*}

Figure \ref{fig:concentric_circles_accuracy} shows the results across a range of $m$ and $\delta$. Figure \ref{fig:concentric_circles_visualization} shows the accuracy achieved by performing spectral clustering on the corresponding similarity graphs. Based on these results, it appears that our clustering algorithm is capable of distinguishing between clusters of arbitrarily small separation $\delta$, provided that the number of atoms is sufficiently large.

\subsection{Real-World Data}

In this section, we empirically evaluate our algorithm on synthetic and three publicly available real-world data sets.  We compared our method against four baseline clustering algorithms that may be interpreted as dictionary learning: (i) $k$-means (KM) \citep{lloyd1982least}, which learns a single dictionary atom for each cluster; (ii) SMCE, which solves a sparse optimization problem over a global dictionary consisting of all data points, then runs spectral clustering on a similarity graph derived from the solution \citep{elhamifar2011sparse}; (iii) LLL \citep{vladymyrov2013locally} which is a landmark method that uses uniform sampling (LLL-U) or k-means clustering (LLL-K) and (iv) 
ESC \citep{you2018scalable} is a landmark method that uses furthest first search. A summary of results can be found in Table \ref{tab:acc}. For LLL, clustering is based on an affinity matrix built from the weights. The reported accuracy is the best result after optimizing for different factors (\# of neighbors, exemplar scheme, optimal clustering). See the Supplementary Materials for details of numerical experiments. We note that the linear system utilized in LLL is ill-conditioned, due to few sample points to characterize the manifold, for the Yale B dataset and obtains poor results. We denote this result by NA. Similarly, clustering on MNIST is ill-conditioned with 500 points and we instead set $m=800$.

\begin{table}[h!]
   \caption{Clustering accuracies for various data sets rounded to three digits. }
    \centering
    \begin{tabular}{|c|c|c|c|c|}
    \hline
        Method & Moons & MNIST-$5$ & Yale B  & Salinas-A \\
        \hline
        KM       & 0.756 & 0.887            &0.508 & 0.774 \\
        SMCE   & 0.835   & 0.975      &1.0 & 0.847 \\
        KDS     &    \textbf{0.999} & \textbf{0.986}            &\textbf{1.0} & \textbf{0.881} \\
        LLL-U     &    0.944 & 0.976            &NA & 0.285 \\
        LLL-K     &    0.950 & 0.980           &NA & 0.261 \\
        ESC           &  0.842           & 0.966                 &  0.958      & 0.840        \\
        \hline
    \end{tabular}
    \label{tab:acc}
\end{table}
\textbf{MNIST Handwritten Digit Database:} The database \citep{lecun1998gradient} consists of $28\times 28$ grayscale images of $10$ different digits. We ran our clustering on a subset of the data comprised of the $k=5$ digits $\{0, 3, 4, 6, 7\}$, following the example of \citep{elhamifar2011sparse}. Figure \ref{fig:training_moon_mnist} shows a subset of the randomly initialized atoms for MNIST before training (black and white) and after training and clustering (color). 

\textbf{Extended Yale Face Database B}: The cropped version of the database \citep{lee2005acquiring} consists of $192 \times 168$ grayscale images of 39 different faces under varying illumination conditions. We
ran our algorithm on a subset of the data comprised of $k = 2$ subjects. 

\begin{figure}[h!]
    \centering
    \includegraphics[width=0.8\linewidth]{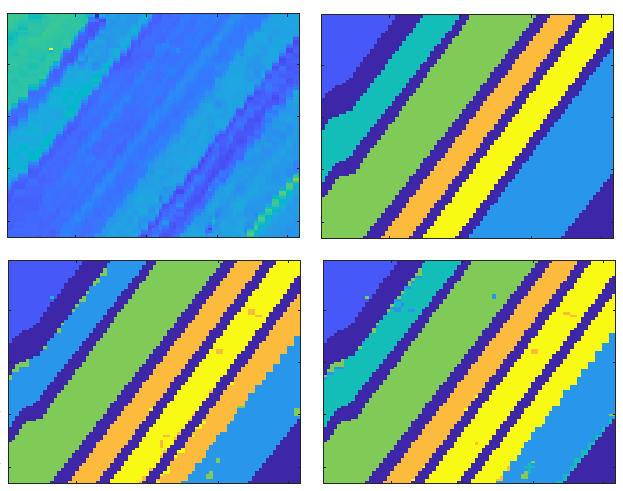}
    \caption{Salinas-A Scene. From left to right and top to bottom: image data (mean across spectral bands), ground truth clusters, predicted clusters by K-means, predicted clusters by KDS.}
    \label{fig:kds_salinas_results}
\end{figure}

\textbf{Salinas-A Hyperspectral Image:} The Salinas-A data set is a single aerial-view hyperpspectral image of the Salinas valley in California with $224$ bands and $6$ regions corresponding to different crops \citep{salinas}. We ran our algorithm on the entire $86\times 83$ pixel image with $k=6$ segments. Regarding the KDS result depicted in Figure \ref{fig:kds_salinas_results}, KDS exhibits a specific limitation: it tends to blend certain elements of the aquamarine class with the yellow class, a characteristic shared with many hyperspectral image (HSI) clustering algorithms. Conversely, K-means exhibits a distinct challenge as it fails not only to distinguish the turquoise class but also struggles to accurately separate a portion of the aquamarine class.

\section{Conclusion}
In this paper, we proposed a structured dictionary learning algorithm $\mathrm{K}$-Deep Simplex (KDS) that combines nonlinear dimensionality reduction and sparse coding. Given a set of data points as an input, KDS learns  a dictionary along with sparse coefficients supported on the probability simplex. Assuming that data points are generated from a convex combination of atoms, represented as vertices of a unique Delaunay triangulation, we prove that the proposed regularization recovers the underlying sparse solution. Furthermore, we demonstrate that when a data point undergoes perturbation and the perturbed point resides within the same d-simplex as the original point, we establish the stability of sparse representations. We also show how the  optimization problem for KDS can be recast and solved via a structured deep autoencoder. 
We then discuss how KDS can be applied for the clustering problem by constructing a similarity graph based on the obtained representation coefficients. Our experiments show that KDS learns meaningful representation and obtains competitive results while offering dramatic savings in running time. In contrast to methods that set the dictionary to be the set of all data points, KDS is quasilinear with the number of dictionary atoms and offers a scalable framework. In our future work, we intend to explore several aspects, including stability estimates for scenarios where perturbed and original data points are located in adjacent d-simplices, conducting experiments on large, real-world datasets, 
examining the sampling of data manifold using KDS and drawing comparisons to \citep{silva2002global,de2004sparse},  investigating the out-of-sample extension property of KDS, and exploring the generative capabilities of the model.\\*

\textbf{Acknowledgements}: AT acknowledges support from NSF through grant DMS-2208392. JMM gratefully acknowledges support from the NSF through grants DMS-1912737, DMS-1924513, DMS-2309519, and DMS-2318894 as well as The Camille \& Henry Dreyfus Foundation.
DB acknowledges support from the NSF through grants DMS-2134157 and PHY-2019786.

\bibliographystyle{IEEEtranN}
\bibliography{paper_references}

\newpage
\appendix
\section{Theoretical analysis of using sparse coefficients for spectral clustering}
In what follows, we provide theoretical guarantees of the proposed framework for the clustering task under some assumptions on the data model.

\subsection{Generative model}

We consider $m$ landmark points $\a_1,\a_2,...,\a_m$ with a unique Delaunay triangulation. In this setting, each point in the set $\{\y_i\}_{i=1}^{n}\in \real^{d}$ is generated from a convex combination of at most $d+1$ atoms.  For simplicity of presentation, we assume there are two clusters. The notion of connectedness is important to analysis of clustering guarantees and we state the following definition in the context of the Delaunay triangulation.

\begin{definition}
Given a set of $n$ points $\{q_j\}_{j=1}^{n}$, we define a graph using an adjacency matrix as follows: $A_{ij} = 1$ if $\q_i$ and $\q_j$ lie in the same or adjacent Delaunay triangles and $A_{ij}=0$
otherwise. We call the set of points  Delaunay-connected if the induced graph is path connected. 
\end{definition}

\begin{figure}[h!]
\begin{center}
    \includegraphics[scale=0.8]{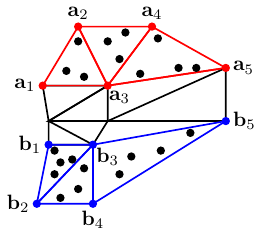}
    \caption{The red dots and blue dots indicate the atoms which define the first and second clusters respectively. Each black dot, denoting a data point, is a convex combination of
    three atoms which are vertices of the triangle the point belongs to. In this case, we see that each cluster is Delaunay-connected and there is no path between two points in different clusters.  }
    \label{fig:cluster_illustration}
    \end{center}
\end{figure}

Figure \ref{fig:cluster_illustration} shows an example with $d=2$ i.e., data points in $\real^{2}$ which belong to two clusters and the set of points with each cluster are Delaunay-connected.  

Given the $m$ landmark atoms, each data point in a triangle can be exactly expressed as a linear combination of the vertices of the triangle. We call the clustering exact if spectral clustering run on the similarity graph of these coefficients identifies the underlying clusters. A simple criterion to ensure that the clustering is exact is that the induced graphs of the two clusters are not Delaunay-connected i.e., there is no path between two points in different clusters.

\begin{theorem}
\label{thm:d_connected}
Consider a set of n points $\{\y_i\}_{i=1}^{n}$ generated from a convex combination of at most $(d+1)$-sparse atoms corresponding to the vertices of a Delaunay triangulation. Assume that there is an underlying clustering of the points into $C_1, C_{2}$.  Define the minimum separation distance between the clusters as $\Delta = \underset{i \in C_1,j \in C_2}{\min}\,||\y_i-\y_j||$. Let $R$ be the maximum diameter of a triangle with the maximum taken among triangles that contain at least one point. If each cluster is Delaunay-connected and $\Delta>2R$, spectral clustering identifies the two clusters exactly. 

\end{theorem}

\begin{proof}[Proof of Theorem \ref{thm:d_connected}]
$\Delta > 2R$ ensures that no triangle contains two points from different clusters. Using this and the fact that each of the clusters are Delaunay-connected, we conclude that spectral clustering
identifies the clusters exactly.  

\end{proof}

We now consider a general model where the points are not necessarily exactly generated from the landmark points.  Suppose each point in the first cluster is noisely generated using the dictionary $\A = [\a_1,...,\a_m]$ and each point in the second cluster is noisely generated using the dictionary $\B = [\b_1,...,\b_p]$. Let $\mathbf{D} = [\A \,\,\B]$.  The points
$[\a_1,...,\a_m]$ and $ [\b_1,...,\b_p]$ are assumed to have a unique Delaunay triangulation. Consider a data point $\y = \sum_{j} x_j\a_j + \eta$ where $\eta$ is additive noise. 
Assume $\epsilon$-closeness of the data point $\y$ in the dictionary $\A$ i.e., there exists representation coefficient $\x$ such that $||\y-\sum_{j}x_j\a_j||\le \epsilon$.
Further, assume also $\epsilon$-closeness of the data point $\y$ in the dictionary $\A'$ where $\A'\subset \A \cup \B$. 
Of interest is the sense in which  $\epsilon$-closeness of the data point $\y$ in the dictionary $\A$ is optimal. To this end, we study the following optimization program
 \begin{align}
&\underset{\x \in \real^{m+p}}{\min} \, \sum_{j} x_j ||\y - \d_j||^2\label{eq:opt_program} \\
& \text{ s.t.}\,\,  ||\y - \sum_{j}x_j\d_j||\le \epsilon\,\,, \,\, \x\ge \mathbf{0} \text{ and } \x\mathbf{1}=\mathbf{1}  \nonumber,
\end{align}
where $\d_j$ denotes the j-th column of the dictionary $\mathbf{D}$. We now define the following two quantities central to the main result to be stated below:
\[
\Delta_1 = \max_{j} ||\y-\a_j||^2, \quad \, \quad  \Delta_2 =\min_{j} ||\y-\a'_j||^2.
\]
\begin{theorem}
\label{thm:delta2_delta1}
If $\Delta_2>\Delta_1$, the optimal solution 
to \eqref{eq:opt_program} is such that it is nonzero only on indices corresponding to $\A$.
\end{theorem}

\begin{proof}[Proof of Theorem \ref{thm:delta2_delta1}]
Assume an $\epsilon$-close reconstruction of a point $\y$ using the dictionary $\A'$. We consider the objective in \eqref{eq:opt_program}
\[
\sum_{j} x_j ||\y-\a'_j||^2 \ge \min_{j} ||\y-\a'_j||^2  \sum_{j} x_j = \Delta_2 \cdot 1 = \Delta_2.
\]
We now consider  $\epsilon$-close reconstruction of a point $\y$ using the dictionary $\A$. We upper bound the objective in \eqref{eq:opt_program}
\[
\sum_{j} x_j ||\y-\a_j||^2 \le \max_{j} ||\y-\a_j||^2 \sum_{j}x_j \le \Delta_1 \cdot 1 = \Delta_1.
\]
Since $\Delta_2> \Delta_1$, $\sum_{j} x_j ||\y-\a_j||^2  < \sum_{j} x_j ||\y-\a'_j||^2$. Therefore, the optimal solution to the optimization program in \eqref{eq:opt_program} is such that $\y$ is $\epsilon$-close in the dictionary $\A$. 

\end{proof}

\textbf{Remark:} The condition that $\Delta_2> \Delta_1$ limits the type of cluster geometries that can be considered in the model.  In the case that each cluster is densely sampled, the noise is small and the clusters are not wide relative to their separation, we expect $\Delta_2> \Delta_1$ to hold.

\section{Details of numerical experiments}

\subsection{Pre-processing of data}
We have conducted pre-processing on the input data using three distinct methods. The first method involves scaling the data to fit within the range of [0,1]. The second method entails standardizing the data to have a mean of 0 and a standard deviation of 1. The third approach focuses on ensuring that each data point possesses a unit norm. When presenting the results for all the methods, we have employed these three pre-processing techniques and reported the best outcome. Here on, $P_1$, $P_2$ and $P_3$ respectively denote the scaling to $[0,1]$, standardizing and normalizing pre-processing of the data.

\subsection{KDS}
For each dataset (Moons, MNIST-5, Yale B, Salinas-A), we trained KDS by backpropagation using the Adam optimizer for a fixed number of epochs. For all experiments, hyperparameters for KDS were chosen using an informal search of the parameter space with the goal of roughly balancing the two terms in the weighted $\ell_1$-regularized loss function. The crucial parameters are $T$ (the number of sparse coding iterations), $\lambda$ which controls the locality regularization, step size for gradient descent, number of epochs and batch size. The optimal network parameters are summarized below.

\begin{table}[htbp!]
    \centering
    \begin{tabular}{|c|c|c|c|c|c|}
    \hline
                      & T & $\lambda$ & Step size  & Epochs & Batch size \\
        \hline
        \vspace{0.2em}
        Moons    & $15$ & $5.0$  & $10^{-3}$ & $10^{3}$  & $10^4$\\
        MNIST-5   & $100$  & 0.5& $10^{-3}$   & $30$  &  $1024$  \\ 
        Yale B   & $50$  & $0.1$  & $10^{-4}$  & $15$ & $1$ \\ 
        Salinas-A   & $100$  & $1$    & $10^{-4}$  & $50$ & $128$\\ \hline
    \end{tabular}
    \caption{Optimal network parameters for KDS}
    \label{tab:kds_optimal}
\end{table}
Table \ref{tab:number of atoms kds} lists the total number of data points and the number of
atoms $m$ used in KDS.
\begin{table}[htbp!]
    \centering
    \begin{tabular}{|c|c|c|c|c|}
    \hline
                      & Moons & MNIST-$5$ & Yale B  & Salinas-A \\
        \hline
        $n$       & 5000 & 5000          & 128  &7138 \\
        $m$   & 24  & 500       & 64       & 25 \\ \hline
    \end{tabular}
    \caption{The number of data points and number of atoms for different datasets used in KDS. }
    \label{tab:number of atoms kds}
\end{table}

\subsection{SMCE}
SMCE depends on a regularization parameter $\lambda$ that controls the sparsity of the representation coefficients. For all experiments, we considered $\lambda \in [1,10,100,200]$ and report the best results. We used the implementation provided by the original authors. Another important parameter is \verb|KMax| which is the number of neighbouring points that constitute a dictionary. To compare SMCE and KDS on the different datasets, we fix the number of dictionary atoms $m$. Table \ref{tab:number of atoms smce} lists the total number of data points and the number of atoms $m$ used in our experiments. We note that, while SMCE with a fixed $m$ resembles KDS, there is a notable difference. In SMCE, each optimization problem uses local dictionaries while KDS employs a fixed learned global dictionary. We note that for Salinas-A, if we use $m=25$ atoms for SMCE, the clustering accuracy is low for SMCE. For that reason, we also report the SMCE accuracy for $m=600$. 

\begin{table}[htbp!]
    \centering
    \begin{tabular}{|c|c|c|c|c|}
    \hline
                      & Moons & MNIST-$5$ & Yale B  & SalinasA \\
        \hline
        $n$       & 5000 & 5000          & 128  &7138 \\
        $m$   & 24  & 500       & 64       & 25 \\ \hline
    \end{tabular}
    \caption{The number of data points and number of atoms for different datasets used in SMCE. }
    \label{tab:number of atoms smce}
\end{table}

\begin{table}[htbp!]
    \centering
    \begin{tabular}{|c|c|c|c|c|}
    \hline
                      & $\lambda=1$ & $\lambda=10$ & $\lambda=100$  & $\lambda=200$ \\
        \hline
        \vspace{0.2em}
        $P_1$    & 0.7812 & 0.9297  & 0.9687 & 0.8437 \\
        $P_2$   & 0.8047  & 0.9766   & 1  &  1 \\ 
        $P_3$   & 0.8359  & 0.992    & 1  & 1 \\ \hline
    \end{tabular}
    \caption{Clustering accuracy for Yale B database (2 faces) using SMCE with $m=64$. }
    \label{tab:smce_yale}
\end{table}

\begin{table}[htbp!]
    \centering
    \begin{tabular}{|c|c|c|c|c|}
    \hline
                      & $\lambda=1$ & $\lambda=10$ & $\lambda=100$  & $\lambda=200$ \\
        \hline
        \vspace{0.2em}
        $P_1$    & 0.8459 & 0.8439  & 0.6599 & 0.6915 \\
        $P_2$   & 0.8469  & 0.7663   & 0.5206  &  0.6257 \\ 
        $P_3$   & 0.8447  & 0.7264    & 0.6245  & 0.6060 \\ \hline
    \end{tabular}
    \caption{Clustering accuracy for Salinas-A dataset using SMCE with $m=600$. }
    \label{tab:smce_salinas}
\end{table}

\begin{table}[htbp!]
    \centering
    \begin{tabular}{|c|c|c|c|c|}
    \hline
                      & $\lambda=1$ & $\lambda=10$ & $\lambda=100$  & $\lambda=200$ \\
        \hline
        \vspace{0.2em}
        $P_1$    & 0.3371 & 0.3240  & 0.5602 & 0.3912 \\
        $P_2$   & 0.4852  & 0.4471   & 0.6113  &  0.5077 \\ 
        $P_3$   & 0.5697  & 0.4131    & 0.6414  & 0.4854 \\ \hline
    \end{tabular}
    \caption{Clustering accuracy for Salinas-A dataset using SMCE with $m=25$. }
    \label{tab:smce_salinas_m25}
\end{table}

\begin{table}[htbp!]
    \centering
    \begin{tabular}{|c|c|c|c|c|}
    \hline
                      & $\lambda=1$ & $\lambda=10$ & $\lambda=100$  & $\lambda=200$ \\
        \hline
        \vspace{0.2em}
        $P_1$    & 0.9610 & 0.9552  & 0.9379 & 0.9574 \\
        $P_2$   & 0.9683  & 0.9749   & 0.9525  &  0.9633 \\ 
        $P_3$   & 0.9679  & 0.9744    & 0.9538  & 0.9632 \\ \hline
    \end{tabular}
    \caption{Clustering accuracy for MNIST-5 dataset using SMCE with $m=500$.  }
    \label{tab:smce_mnist}
\end{table}

\begin{table}[htbp!]
    \centering
    \begin{tabular}{|c|c|c|c|c|}
    \hline
                      & $\lambda=1$ & $\lambda=10$ & $\lambda=100$  & $\lambda=200$ \\
        \hline
        \vspace{0.2em}
        $P_1$    & 0.8352 & 0.8274  & 0.6664 & 0.5400 \\
        $P_3$   & 0.5562  & 0.5204    & 0.5322  & 0.5284 \\ \hline
    \end{tabular}
    \caption{Clustering accuracy for Moons dataset using SMCE with $m=24$. The pre-processing $P_2$ gives redundant columns which is degenerate as input for SMCE.}
    \label{tab:smce}
\end{table}

\subsection{LLL}

We employed the implementation by the authors of the LLL algorithm. This algorithm employs two distinct approaches for selecting landmarks: LLL-U utilizes uniform sampling, while LLL-K relies on K-means clustering. When applying LLL to the Yale B dataset, which comprises 128 facial images across 2 classes, we noticed through empirical experimentation that the linear system in LLL becomes poorly conditioned due to the limited number of sample points. As a consequence, this adversely impacts the clustering accuracy and the data may not accurately represent the natural setting for this technique. Consequently, we have chosen not to present the results for the Yale B dataset. Another crucial parameter in LLL is denoted as $K$ which represents the number of landmarks that each data point utilizes. Table \ref{tab:number of atoms lll} provides a summary of this parameter, along with the total number of landmarks employed within the LLL method. Given the optimal representation coefficients from LLL, we apply spectral clustering on the coefficients for the moons and Salinas A dataset. In the case of MNIST-5, we construct a similarity matrix by employing the $k$-nearest neighbor graph among data points. Subsequently, we apply the LLL method to obtain optimal embeddings from this similarity matrix. These resultant embeddings are the input to K-means clustering. We have found that the aforementioned configurations and parameter selections, as summarized in Table \ref{tab:number of atoms lll}, consistently yield favorable results  across the parameter ranges we have explored based. This entails careful consideration of factors like selecting the number of landmarks to closely match KDS while mitigating potential ill-conditioning errors in LLL. The clustering accuracy for various datasets can be found in both Table \ref{tab:accuracy lllu} and Table \ref{tab:accuracy lllk}.

\begin{table}[htbp!]
    \centering
    \begin{tabular}{|c|c|c|c|}
    \hline
                      & Moons & MNIST-$5$   & Salinas-A \\
        \hline
                \vspace{0.2em}
        $K$       & 24 & 11            &600 \\
        $m$   & 24  & 800   & 600 \\ \hline
    \end{tabular}
    \caption{Number of landmarks each data point utilizes and total number of landmarks for different datasets used in LLL. }
    \label{tab:number of atoms lll}
\end{table}

\begin{table}[htbp!]
    \centering
    \begin{tabular}{|c|c|c|c|}
    \hline
                      & Salinas-A& MNIST-5 & Moons   \\
        \hline
        \vspace{0.1em}
        $P_1$    & 0.2853 & 0.9696  & 0.9442  \\
        $P_2$   & 0.2152  & 0.9751   & NA   \\ 
        $P_3$   & 0.2121  & 0.9758   & 0.8047  \\ \hline
    \end{tabular}
    \caption{Clustering accuracy for the different datasets using LLL-U. Clustering accuracies are averages of $5$ runs.}
    \label{tab:accuracy lllu}
\end{table}

\begin{table}[htbp!]
    \centering
    \begin{tabular}{|c|c|c|c|}
    \hline
                      & Salinas-A& MNIST-5 & Moons   \\
        \hline
        \vspace{0.1em}
        $P_1$    & 0.2612 & 0.9735  & 0.9504  \\
        $P_2$   & 0.2104  & 0.9797   & NA   \\ 
        $P_3$   & 0.2019  & 0.9790    & 0.8194  \\ \hline
    \end{tabular}
    \caption{Clustering accuracy for the different datasets using LLL-K. Clustering accuracies are averages of $5$ runs.}
    \label{tab:accuracy lllk}
\end{table}

\subsection{ESC}

We employed the implementation by the authors of the ESC algorithm. 
In our numerical experimentation, we note that the quality of the Lasso solver in ESC determines the final clustering accuracy. Given that, we use the default setting which uses a Lasso solver from SPAMS package \citep{mairal2010online}.
Crucial parameters in the ESC algorithms are number of exemplars $(m)$, number of nearest neighbours $(t)$ and penalty parameter $\lambda$ for the underlying Lasso problem. We set $t=3$ and $\lambda=200$ for all our experiments.
 Table \ref{tab:number of atoms esc} provides a summary of these parameters. We have found that these parameter choices, consistently yield favorable results with number of landmarks comparable to KDS. The clustering accuracy for various datasets can be found in Table \ref{tab:accuracy esc}. Since there is variation in the clustering accuracy each run for Yale B and Salinas, we have reported an average result from 5-trials. We note that the first data pre-processing, denoted as $P_1$, consistently triggers an error in the ESC code for the Moons dataset. Consequently, we have recorded this outcome as NA.

\begin{table}[htbp!]
    \centering
    \begin{tabular}{|c|c|c|c|c|}
    \hline
                      & Moons & MNIST-$5$   & Yale B & Salinas-A \\
        \hline
                \vspace{0.2em}
          $m$   & 24  & 500   & 32 & 600 \\ \hline
           $t$   & 48  & 3   & 2 & 3 \\ \hline
    \end{tabular}
    \caption{Number of exemplars ($m)$  and total number of neighbours($t$) for different datasets used in ESC. }
    \label{tab:number of atoms esc}
\end{table}

\begin{table}[htbp!]
    \centering
    \begin{tabular}{|c|c|c|c|c|c|}
    \hline
                      & Yale B & Salinas-A& MNIST-5 & Moons   \\
        \hline
        \vspace{0.1em}
        $P_1$    &0.6750  & 0.7774 & 0.5661   & NA  \\
        $P_2$ &0.8953  & 0.7137  & 0.6168   & 0.5030   \\ 
        $P_3$ &0.9578  & 0.8403  & 0.9658 & 0.8426  \\ \hline
    \end{tabular}
    \caption{Clustering accuracy for the different datasets using ESC. }
    \label{tab:accuracy esc}
\end{table}

\subsection{Choosing $m$}

The number of dictionary atoms $m$ is a central parameter of KDS.  Because the dictionary is global, rather than local, it does not scale with the dimension of the data only.  Indeed, $m$ must be large enough to ensure that \emph{any} observed data point is well-approximated by a sparse combination of the atoms.  However, under the model that the data are sampled from a mixture of $K$ probability measures supported on $d$-dimensional manifolds, $m$ simply needs to be chosen large enough to provide an $\epsilon$ covering of the data.  This can be done \citep{gyorfi2006distribution} taking $m=C\epsilon^{-\frac{1}{d}}\log(\epsilon^{-\frac{1}{d}})$, where $C$ is a constant depending on the geometric properties of the underlying manifolds (e.g., their curvatures).  Importantly, $m$ can be taken independently from $n$ and is ``cursed" only by the intrinsic dimensionality of the data.  This is most interesting in the case when $n$ is large.  Indeed, if $n$ is small, then $m=n$ is computationally tractable and taking a dictionary consisting of all observed data points will be optimal; this essentially reduces to the SMCE formulation.  

Together, this suggests $m\ll n$ is possible for KDS while still achieving excellent performance, particularly when $n$ is large.  Our experimental results bear this out, especially for datasets with known low-dimensional structure.  In the two moons data, for example, we take $m=24$ even when $n=5000$.  This is possible because the intrinsic dimensionality of each moon is 1 in the noiseless case and approximately 1 in the presence of low-variance Gaussian noise. 

In addition, in our numerical experiments, we observe that choosing large values of $m$ could lead to sub-optimal results. The primary observation pertains to the fact that the sparse representation coefficients are highly disconnected i.e., the data points do not share sufficiently many atoms for representation. While a certain degree of disconnectedness is desirable for spectral clustering, there is no precise way to cluster $p$ disconnected groups when $p$ exceeds the number of clusters. With that, $m$ should be set suitably small and increased progressively or set by methods such as cross validation by looking at the optima loss values of the optimization objective.

\section{Computational Complexity}
The original implementation of SMCE is in MATLAB and the code is available at \url{http://vision.jhu.edu/code/}.
KDS is implemented in the PyTorch framework \citep{paszke2019pytorch}. Given these differences and the choice of different algorithms or routines for the optimization programs and solvers, elapsed times do not fairly provide conclusive evidences for the computational advantages of one method over another. Given this, we focus on showing how computational time scales with $n$. 
\subsection{Complexity of SMCE}
\begin{figure*}[htbp!]
    \centering
    \includegraphics[width=0.9\linewidth]{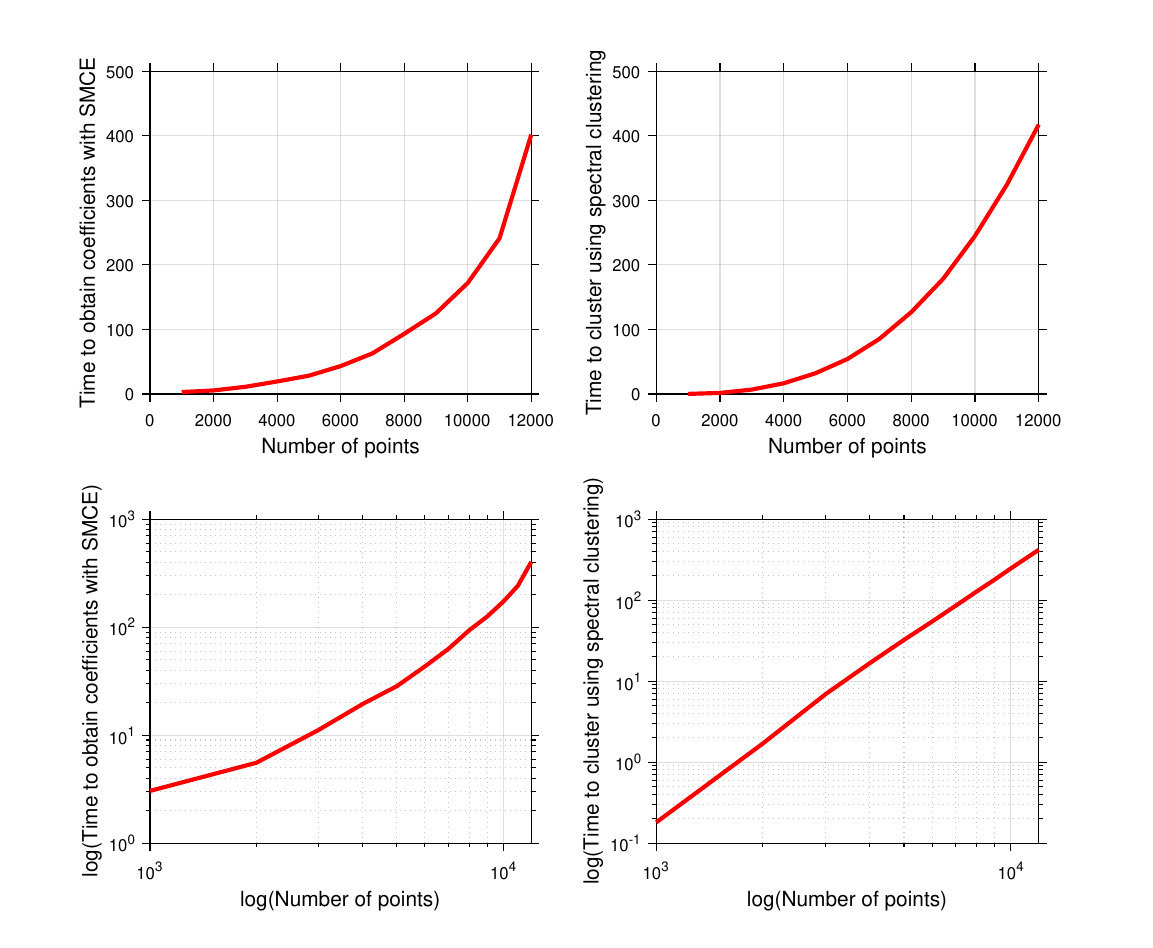}
    \caption{(Left) Number of data points vs time taken to obtain coefficients for the two moon dataset, (Right) Number of data points vs time taken to cluster the two moon dataset based on a similarity graph constructed from the coefficients. The top two plots are on the standard scale and the bottom two plots are on a log-log scale. As $n$ increases, the cost of SMCE becomes prohibitive and motivates a scalable method like KDS. As we discuss in the next section, KDS benefits from linear scaling in the number of data points for both tasks of obtaining coefficients and spectral clustering. }
    \label{fig:SMCE_complexity_plots}
\end{figure*}

We consider different number of data points from the two moons data corrupted with small Gaussian noise. We set the number of dictionary atoms $m$ to be $n/10$ where $n$ is
the total number of data points. This is the setting used in all the numerical experiments in \citep{elhamifar2011sparse}. Table \ref{tab:SMCE_complexity} shows the time
SMCE takes to obtain coefficients for all data points and the time it takes to cluster the two moon data as the $n$ varies from $1000$ to $12000$. Figures \ref{fig:SMCE_complexity_plots}
show how these times scale with $n$ on the standard and log-log scales. A linear fit of the graphs on the log-log scale gives slopes $1.97$ and $3.1$ respectively. This suggests that obtaining coefficients for SMCE has complexity $O(n^2)$ and spectral clustering on the coefficients costs $O(n^3)$. 
\begin{table}[htbp]
    \centering
    \begin{tabular}{|c|c|c|}
    \hline
       Number of data points    & $t_1$ (sec) & $t_2$ (sec)  \\         \hline
               1000 & 3.04 & 0.18\\
               2000 & 5.54 & 1.70\\
               3000 & 11.15 & 6.86\\
               4000 & 19.40 & 16.61\\
               5000 & 28.39 & 32.23\\
               6000 & 43.22 & 54.25\\
               7000 & 62.93 & 85.34\\
               8000 & 93.19 & 126.78\\
               9000 & 125.02 & 178.40\\
               10000 & 171.90 & 244.93\\
               11000 & 241.11 & 324.31\\
               12000 & 401.77  & 417.40        \\      \hline
       \end{tabular}
    \caption{($t_1$) Time SMCE takes to obtain coefficients, ($t_2$) Time it takes to cluster the noisy two moon dataset.  }
    \label{tab:SMCE_complexity}
\end{table}

\subsection{Complexity of KDS}
\begin{figure*}
    \centering
    \includegraphics[width=0.9\linewidth]{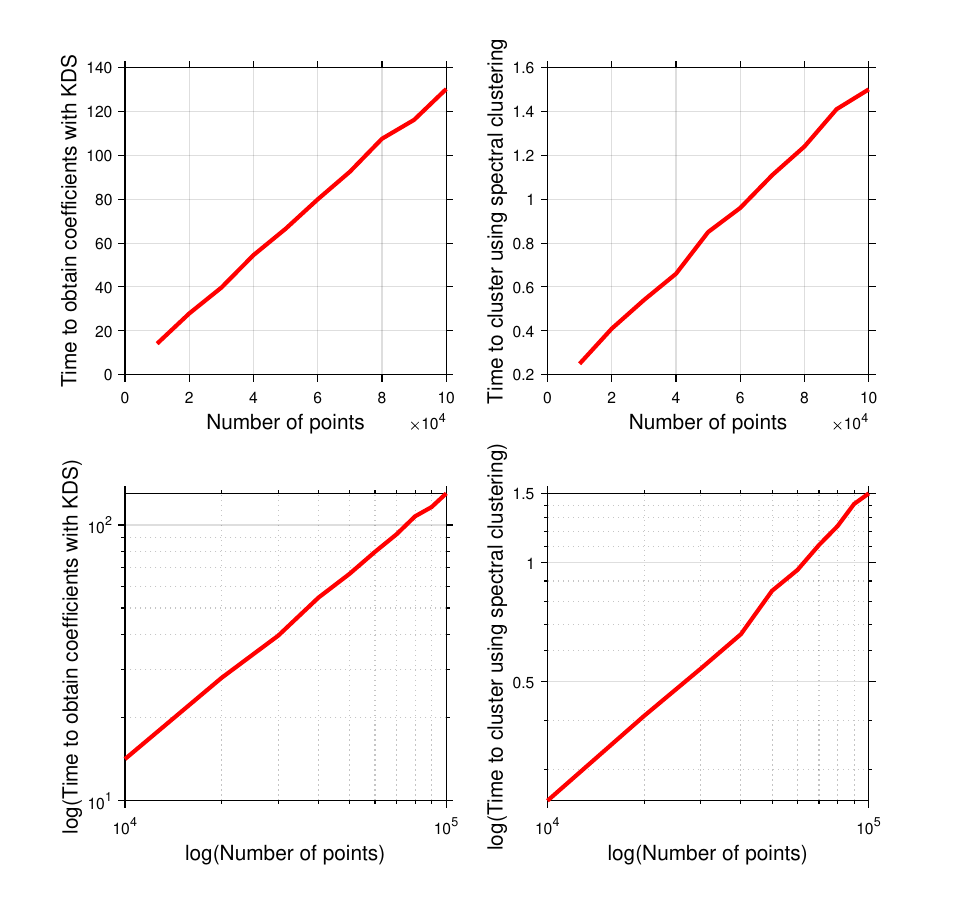}
    \caption{(Left) Number of data points vs time taken to obtain coefficients for the two moon dataset, (Right) Number of data points vs time taken to cluster the two moon dataset based on a similarity graph constructed from the coefficients. The top two plots are on the standard scale and the bottom two plots are on a log-log scale.  }
    \label{fig:KDS_complexity_plots}
\end{figure*}

We consider different number of data points from the two moons data corrupted with small Gaussian noise. We set the number of dictionary atoms $m$ to be $24$.
Table \ref{tab:KDS_complexity} shows the time SMCE takes to obtain coefficients for all data points and the time it takes to cluster the two moon data as the $n$ varies from $10000$ to $100000$. 
Two remarks are in order. First, in the regime of high number of data points, analogous experiments for SMCE do not complete on a standard laptop. Second, even with $m=24$
dictionary atoms, the clustering accuracy is at its minimum $97\%$. Figure \ref{fig:KDS_complexity_plots} shows how these times scale with $n$ on the standard and log-log scales. A linear fit of the graphs on the log-log scale gives slopes $0.97$ and $0.79$ respectively. This suggests that obtaining coefficients for
KDS has complexity $O(n)$ and the spectral clustering step costs at most $O(n)$. 
\begin{table}[htbp!]
    \centering
    \begin{tabular}{|c|c|c|}
    \hline
       Number of data points    & $t_1$ (sec) & $t_2$ (sec)  \\         \hline
               10000 & 14.16 & 0.25\\
               20000 & 27.92 & 0.41\\
               30000 & 39.7 & 0.54\\
               40000 & 54.5 & 0.66\\
               50000 & 66.5 & 0.85\\
               60000 & 79.91 & 0.96\\
               70000 & 92.49 & 1.11\\
               80000 & 107.52 & 1.24\\
               90000 & 116.13 & 1.41\\
               100000 & 130.18 & 1.5\\ \hline      
       \end{tabular}
    \caption{($t_1$) Time KDS takes to obtain coefficients, ($t_2$) Time it takes to cluster the noisy two moon dataset.}
    \label{tab:KDS_complexity}
\end{table}

\section{Numerical exploration of $\sigma_{\min}(\B_L)$}

The main stability result of this manuscript, stated in Theorem 3, depends on the level of noise and  $\sigma_{\min}(\B_L)$. We recall the definition of $\B_L$ as provided in Definition 8. Given a set of $(d+1)$ points in $\real^d$ that define a local dictionary $\A_L \in \real^{d\times(d+1)}$, $\B_L$ is constructed by appending a row vector of ones to $\A_L$. As the stability hinges on the minimum singular value of $\B_L$, an important question concerns the interplay between the configuration of the local dictionary and the value of the minimum singular value. While a comprehensive analysis of this connection has not been undertaken within the current manuscript, we present experimental findings that offer a partial understanding of how the minimum singular value relates to the geometry of the local dictionary.

\begin{figure*}[tbp]
    \centering
    \includegraphics[width=1\linewidth]{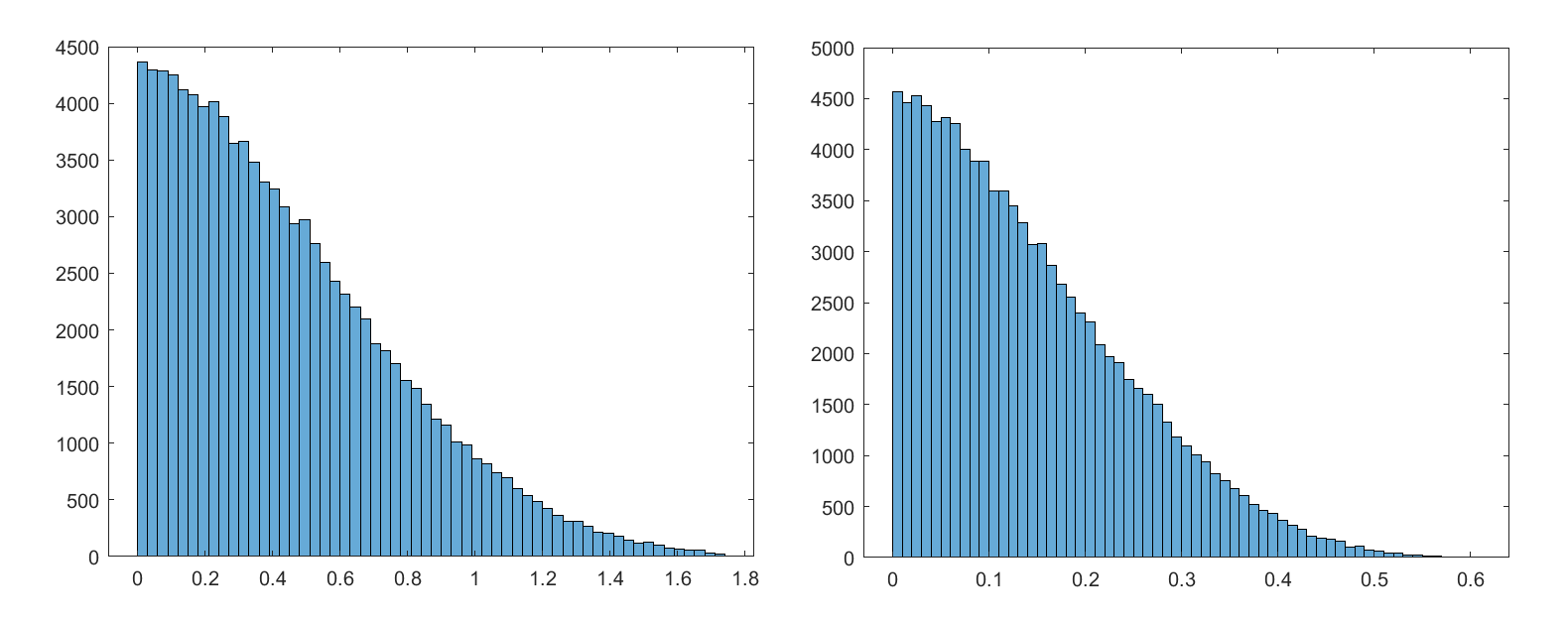}
    \caption{(Left) Histogram of the minimum singular value of $\B_L$ where points are realized from a normal distribution. (Right) Histogram of the minimum singular value of $\B_L$ where points are realized from a uniform distribution. The experiments are repeated $10^5$ times.}
    \label{fig:singular_value_min}
\end{figure*}

For the numerical experiment, we generate $3$ points in $\real^2$ whose coordinates are sampled from the uniform or normal distribution. For each realization, we construct a local dictionary and proceed to compute the minimum singular value of $\B_L$. This process is repeated $10^5$ times. Figure \ref{fig:singular_value_min} is a histogram that shows the distribution of of the minimum singular values. Out of all the trials, we also identify the local dictionaries that correspond to the minimum and maximum $\sigma_{\min}(\B_L)$. For the uniform distribution, the following are the local dictionaries corresponding to the minimum and maximum $\sigma_{\min}(\B_L)$:
\begin{align*}
(\A_L)_{\min} &=
\begin{pmatrix}
   0.3557  &  0.3472 &    0.5490\\
    0.3313 &   0.3242  &  0.4931\\  
\end{pmatrix}\\
   (\A_L)_{\max} &=
\begin{pmatrix}
     0.0369  &   0.9998  &    0.2929\\
    0.9289  &   0.9399  &  0.0148\\  
\end{pmatrix}.
\end{align*}
For the normal distribution, the local dictionaries corresponding to the minimum and maximum $\sigma_{\min}(\B_L)$ are given below
\begin{align*}
(\A_L)_{\min} &=
 \begin{pmatrix}
 1.3696  & -0.3179  &    0.2321\\
  -0.0483 &   0.5274  &   0.3397     
 \end{pmatrix}\\
(\A_L){\max} &=
\begin{pmatrix}
   1.0275  & -1.4530  &   0.4005\\
   -1.5784 &  -0.1859 &   1.7665\\
\end{pmatrix}.
  \end{align*}

\begin{figure*}[htbp!]
    \centering
    \includegraphics[scale=0.9]{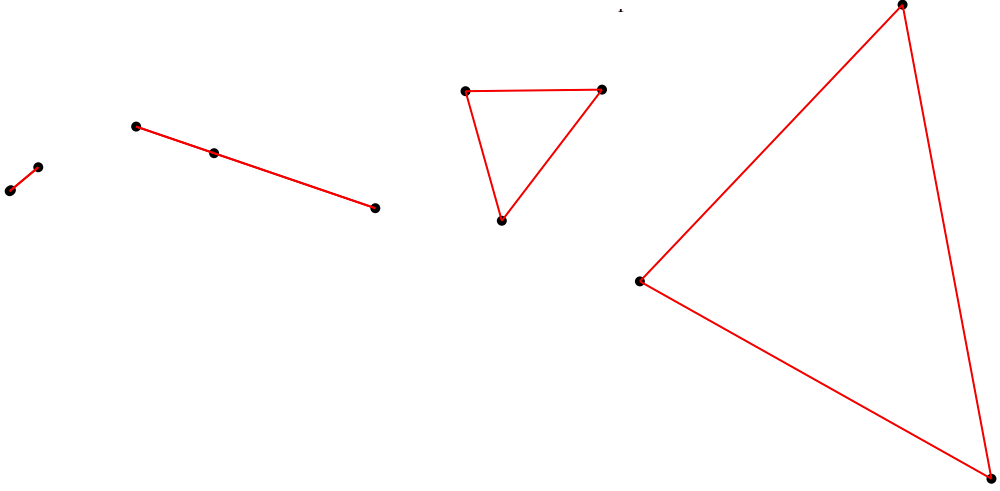}
    \caption{(Left to right) The first two figures show the local dictionaries corresponding to the lowest $\sigma_{\min}(\B_L)$ in a trial of $10^5$ using uniform and normal distribution respectively. The last two figures show the local dictionaries corresponding to the highest $\sigma_{\min}(\B_L)$ in a trial of $10^5$ using uniform and normal distribution respectively. }%
    \label{fig:sigma_min_visualize}%
\end{figure*}

 The visual representations of the aforementioned localized dictionaries are presented in Figure \ref{fig:sigma_min_visualize}. It is evident that the minimum $\sigma_{\min}(\B_L)$ corresponds to \emph{degenerate points} i.e., points that are nearly collinear. Conversely, the maximum $\sigma_{\max}(\B_L)$ correlates with triangles exhibiting a ``well-structured" configuration. In our numerical trials, it is worth mentioning that the visualizations associated with small $\sigma_{\min}(\B_L)$ are elongated triangles.

\section{Visualizing the embeddings of KDS}

\begin{figure*}[tbp]
    \centering    \includegraphics[scale=0.9]{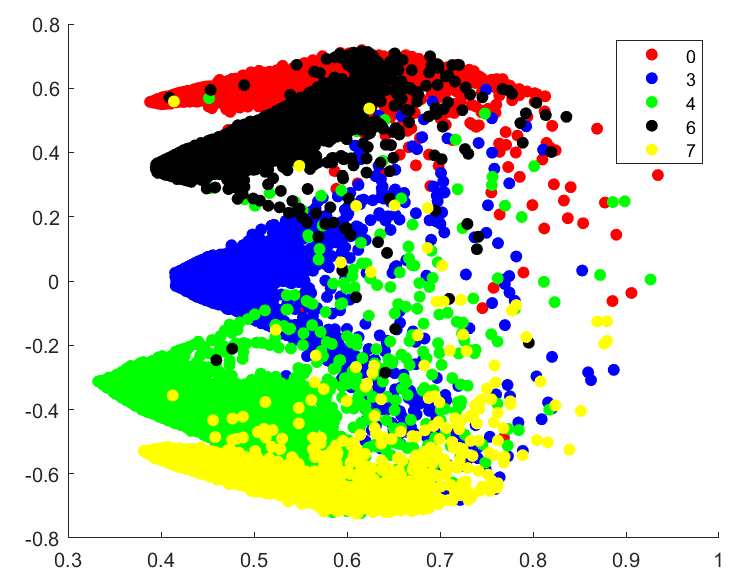}
    \caption{The first 2 coordinates of the KDS embedding. The color codings are according to the ground truth label. }
    \label{fig:mnist2d}
\end{figure*}

\begin{figure*}[tbp]
    \centering    \includegraphics[scale=0.9]{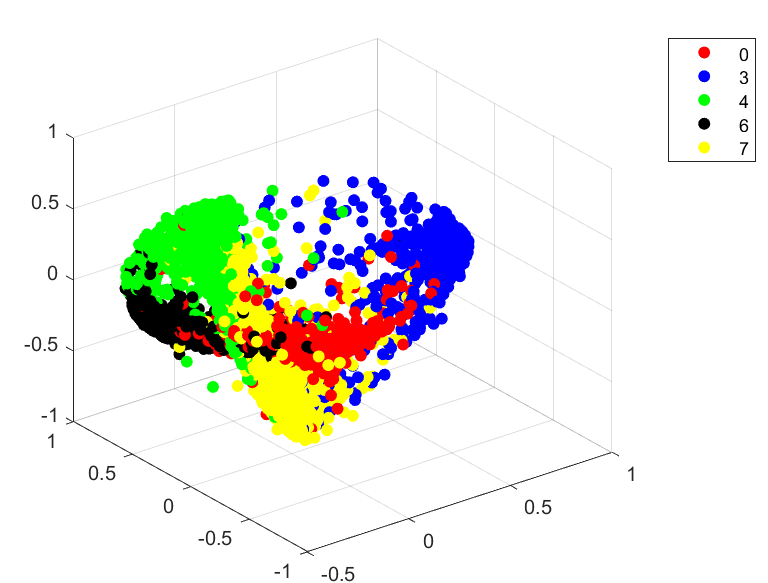}
    \caption{The last 3 coordinates of the KDS embedding. The color codings are according to the ground truth label. }
    \label{fig:mnist3d}
\end{figure*}

In this portion, we provide visual representations of KDS embeddings using MNIST-5 as a demonstrative case. For MNIST-5, these embeddings are in five dimensions. We display the initial two coordinates in two-dimensional space (Figure \ref{fig:mnist2d}), while the subsequent three coordinates are presented in three dimensions (Figure \ref{fig:mnist3d}).

\section{Visualizing the sparse representations of MNIST}

\begin{figure*}[tbp]
    \centering    \includegraphics[scale=0.6]{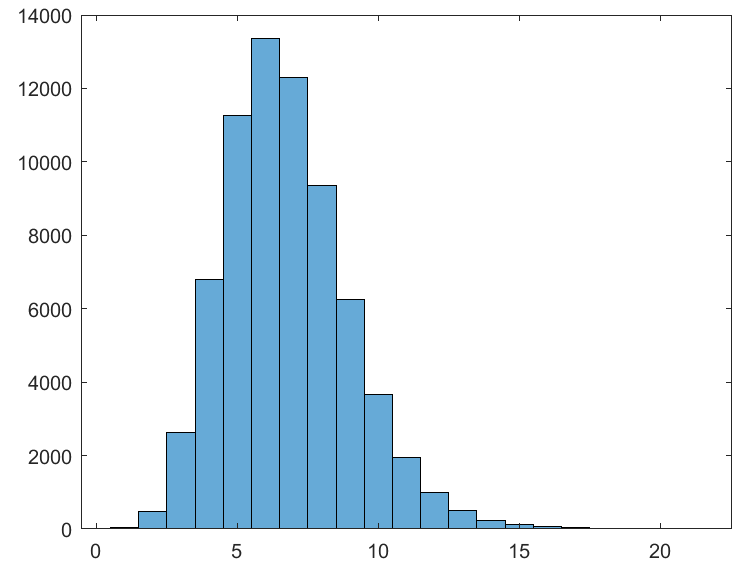}
    \caption{Distribution for the number of atoms used by each MNIST digit. The histogram illustrates that each digit is encoded using only few atoms from the available pool of 1000 atoms. }
    \label{fig:histogram_mnist_kds}
\end{figure*}

\begin{figure*}[tbp]
    \centering    \includegraphics[scale=0.6]{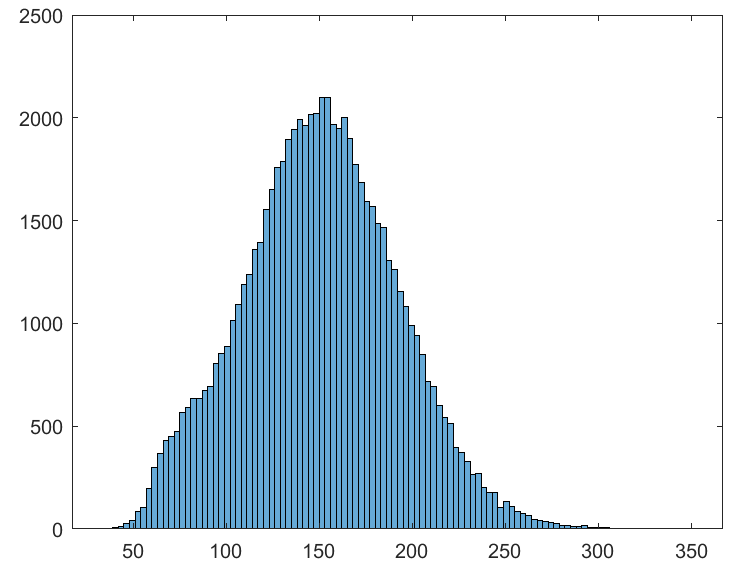}
    \caption{Distribution for the number of non-zero entries in the MNIST database.  }
    \label{fig:histogram_bare}
\end{figure*}

\begin{figure*}
    \centering
    \includegraphics[scale=0.4]{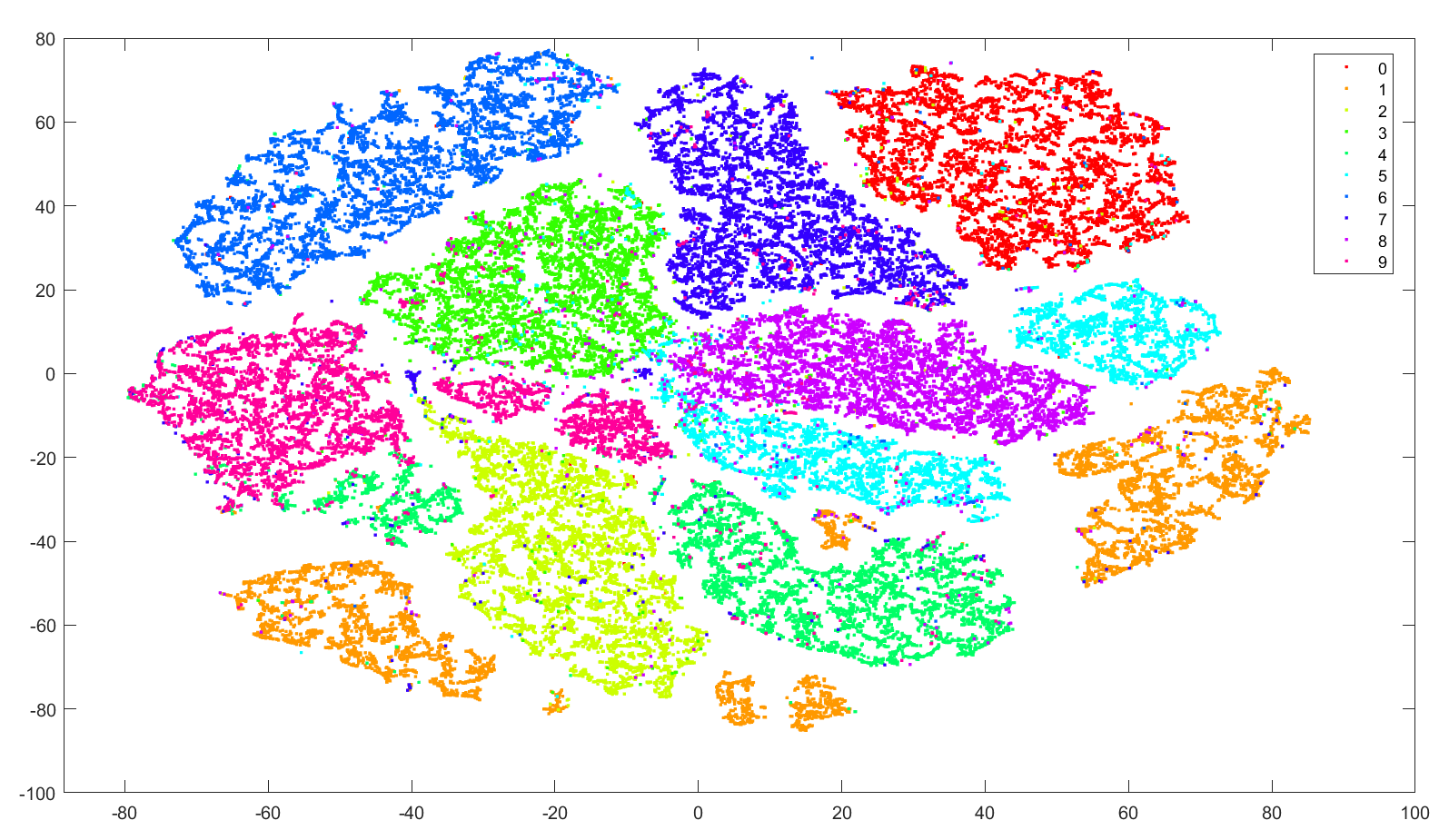}
    \caption{TSNE using KDS coefficients}
    \label{fig:tsne_kds}
\end{figure*}

\begin{figure*}
    \centering
    \includegraphics[scale=0.4]{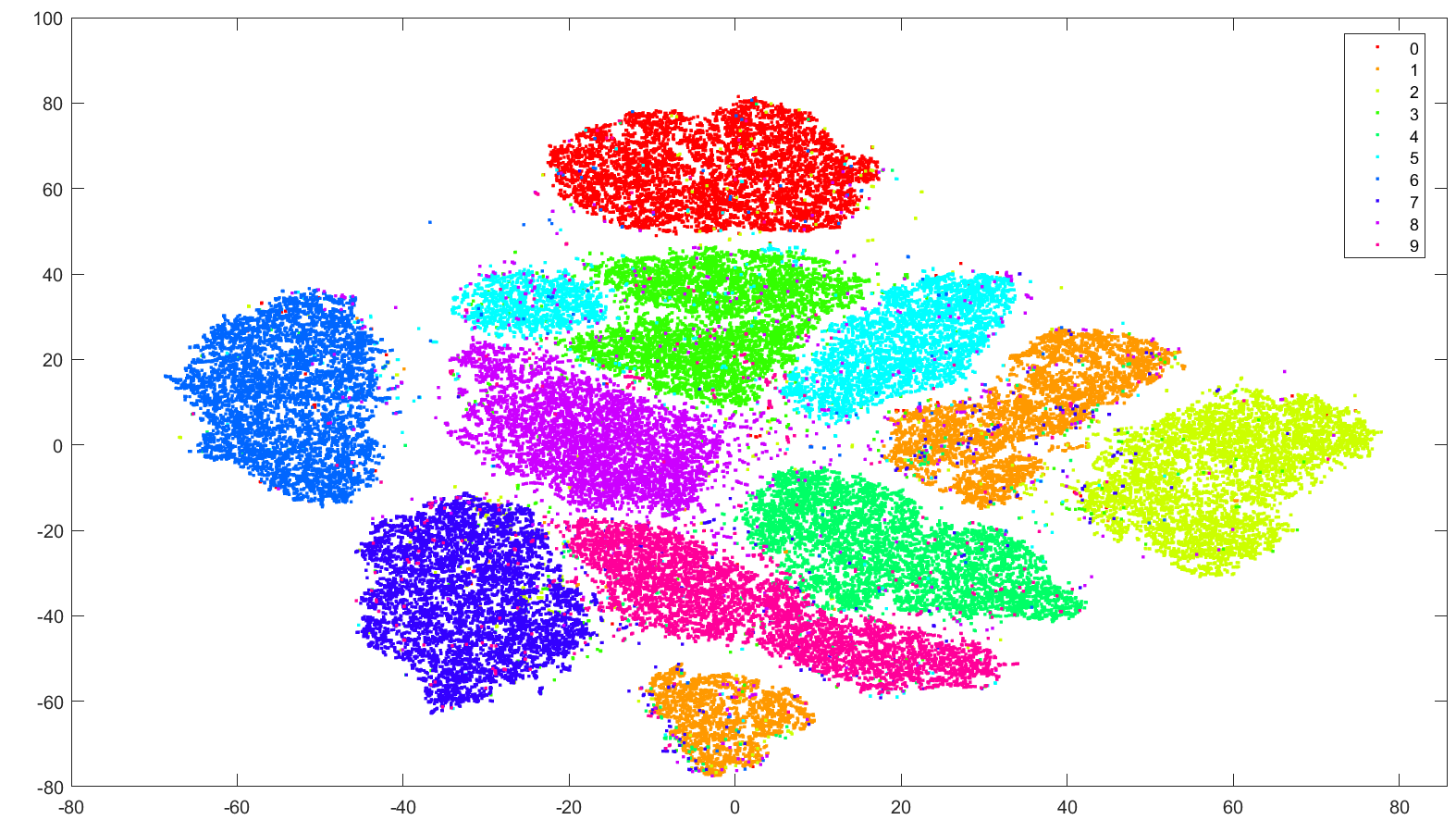}
    \caption{TSNE using original MNIST representation}
    \label{fig:tsne_mnist}
\end{figure*}

In this section, we explore the optimal sparse representations learned by KDS for the MNIST data. We utilize t-Distributed Stochastic Neighbor Embedding (t-SNE) \citep{van2008visualizing} to visually depict the optimal sparse representations derived from our algorithm. In order to establish a point of reference, we also present a t-SNE plot of the unaltered MNIST dataset. The outcomes are showcased in Figure \ref{fig:tsne_kds} and \ref{fig:tsne_mnist}. It's important to highlight that we execute the t-SNE computation using the same metric ($\ell_1$ norm) for both methods, employing identical parameters. Since the application of t-SNE encompasses the entirety of the MNIST dataset, the computation employs the Barnes-hut approximation \citep{van2013barnes}. 
Figure \ref{fig:histogram_mnist_kds} shows a histogram showing the distribution for number of atoms (out of 1000 atoms) used by each individual MNIST digit. Figure \ref{fig:histogram_bare} shows a histogram showing the distribution for the number of non-zero entries in the MNIST database. 

Notably, despite the sparsity of the representations for most digits, the t-SNE representation derived from KDS bears a resemblance to the t-SNE representation of the original digits. This underscores the idea that the sparse representations facilitate downstream tasks while capturing inherent structures within the original data.

\end{document}